%% file: iclr2027_conference.tex
\documentclass{article} 
\usepackage{iclr2027_conference,times}

\input{math_commands.tex}

\usepackage{tabularx}
\usepackage{array}
\usepackage{hyperref}
\usepackage{url}
\usepackage{lipsum}
\usepackage{mwe}
\usepackage{etoc}

\usepackage{wrapfig}
\usepackage{booktabs}
\usepackage{multirow}
\usepackage{subcaption}
\usepackage{cleveref}
\usepackage{amssymb}
\input{styles/prompt_boxes}

\title{Learning to Ask: Information Acquisition for SLM-LLM Collaboration, under a budget}

\author{
Yongjun Kim$^{1}$,
Xiaoxiao Li$^{2,3}$, 
Jaeho Lee$^{1}$ \\
$^{1}$Pohang University of Science and Technology (POSTECH) \\
$^{2}$Department of Electrical and Computer Engineering,
University of British Columbia \\
$^{3}$Vector Institute \\
\texttt{\{yongjun.kim,jaeho.lee\}@postech.ac.kr} \\
\texttt{xiaoxiao.li@ece.ubc.ca}
}

\iclrfinalcopy 
\begin{document}

\maketitle

\begin{abstract}
Collaboration between a small language model (SLM) and a large language model (LLM) offers an opportunity to combine the efficiency of smaller models with the strong reasoning capabilities of larger ones.
Existing approaches primarily frame such collaboration as a computation allocation problem, determining which model should handle each portion of the reasoning process.
In black-box API-based settings, however, this paradigm can be inefficient due to coarse-grained delegation or repeated transmission of context across model switches.
In this work, we instead formulate SLM-LLM collaboration as an \textit{information acquisition} problem, under an API budget constraint.
The SLM remains the primary reasoner and selectively queries a black-box LLM advisor only when needed, issuing targeted queries rather than delegating the reasoning process itself.
To realize this strategy, we develop a three-stage RLVR framework that learns whether to call the advisor, how to formulate useful queries, and how to integrate the collaboration into the reasoning process by jointly refining advisor invocation and information use.
Across mathematical reasoning and coding tasks, our approach improves the performance--cost tradeoff over existing collaboration baselines and, in some settings, matches or exceeds oracle problem-level routing.
Finally, we show that our strategy can transfer to other advisor model families, without further training.
\end{abstract}

\input{section/1_introduction}

\input{section/2_related_work}

\input{section/3_problem_formulation}

\input{section/4_pilot_study}
\input{section/5_method_learning_to_question}
\input{section/6_experiments}

\input{section/7_analysis}

\input{section/8_conclusion}
\input{section/ai_use}

\bibliography{iclr2027_conference}
\bibliographystyle{iclr2027_conference}
\clearpage
\appendix
\addcontentsline{toc}{chapter}{Appendix}
\localtableofcontents
\clearpage
\input{section/9_appendix}

\end{document}

%% file: math_commands.tex
\usepackage{amsmath,amsfonts,bm}

\def\eqref#1{equation~\ref{#1}}

\def\1{\bm{1}}

\DeclareMathAlphabet{\mathsfit}{\encodingdefault}{\sfdefault}{m}{sl}
\SetMathAlphabet{\mathsfit}{bold}{\encodingdefault}{\sfdefault}{bx}{n}



%% file: styles/prompt_boxes.tex
\usepackage[most]{tcolorbox}
\usepackage{listings}
\usepackage{xcolor}

\definecolor{promptgray}{HTML}{B8B8B8}
\definecolor{toolblue}{HTML}{7895B2}
\definecolor{querygreen}{HTML}{7E9F88}
\definecolor{resultamber}{HTML}{B39768}
\definecolor{labelgray}{HTML}{666666}
\definecolor{boxborder}{HTML}{D0D0D0}
\definecolor{boxbackground}{HTML}{FBFBFB}
\definecolor{modeltwo}{HTML}{888888}

\newcommand{\promptlabel}[1]{%
  \par\noindent
  {\small\bfseries\color{labelgray}#1}
  \par\vspace{2pt}
}

\lstdefinestyle{promptstyle}{
  basicstyle=\ttfamily\small,
  breaklines=true,
  breakatwhitespace=true,
  breakindent=0pt,
  columns=fullflexible,
  keepspaces=true,
  showstringspaces=false,
  xleftmargin=0pt,
  framexleftmargin=0pt,
  aboveskip=0pt,
  belowskip=0pt
}

\newtcolorbox{systempromptboxmath}[1][SLM prompt for Math]{
  enhanced,
  breakable,
  colback=boxbackground,
  colframe=boxborder,
  boxrule=0.5pt,
  arc=2pt,
  left=2mm,
  right=2mm,
  top=2mm,
  bottom=2mm,
  before skip=6pt,
  after skip=8pt,
  title={#1},
  colbacktitle=modeltwo,
  coltitle=white,
  fonttitle=\bfseries
}

\newtcolorbox{systempromptboxcode}[1][SLM prompt for Code]{
  enhanced,
  breakable,
  colback=boxbackground,
  colframe=boxborder,
  boxrule=0.5pt,
  arc=2pt,
  left=2mm,
  right=2mm,
  top=2mm,
  bottom=2mm,
  before skip=6pt,
  after skip=8pt,
  title={#1},
  colbacktitle=modeltwo,
  coltitle=white,
  fonttitle=\bfseries
}

\newtcolorbox{systempromptfilter}[1][Prompt for filtering sensitive text]{
  enhanced,
  breakable,
  colback=boxbackground,
  colframe=boxborder,
  boxrule=0.5pt,
  arc=2pt,
  left=2mm,
  right=2mm,
  top=2mm,
  bottom=2mm,
  before skip=6pt,
  after skip=8pt,
  title={#1},
  colbacktitle=modeltwo,
  coltitle=white,
  fonttitle=\bfseries
}

\newtcblisting{promptboxjudge}[1][Prompt for the judge]{
  enhanced,
  breakable,
  listing only,
  listing engine=listings,
  colback=boxbackground,
  colframe=boxborder,
  boxrule=0.5pt,
  arc=2pt,
  left=2mm,
  right=2mm,
  top=2mm,
  bottom=2mm,
  before skip=6pt,
  after skip=6pt,
  title={#1},
  colbacktitle=promptgray,
  coltitle=white,
  fonttitle=\bfseries,
  attach boxed title to top left={
    xshift=0pt,
    yshift=0pt
  },
  boxed title style={
    colframe=promptgray,
    colback=promptgray,
    boxrule=0pt,
    arc=1.5pt,
    left=2mm,
    right=2mm,
    top=1mm,
    bottom=1mm
  },
  listing options={
    basicstyle=\ttfamily\small,
    breaklines=true,
    breakindent=0pt,
    columns=fullflexible,
    keepspaces=true,
    showstringspaces=false,
    xleftmargin=0pt,
    aboveskip=0pt,
    belowskip=0pt
  }
}

\newtcolorbox{promptboxqmaker}[1][Qmaker prompt]{
  enhanced,
  breakable,
  colback=boxbackground,
  colframe=boxborder,
  boxrule=0.5pt,
  arc=2pt,
  left=2mm,
  right=2mm,
  top=2mm,
  bottom=2mm,
  before skip=6pt,
  after skip=8pt,
  title={#1},
  colbacktitle=modeltwo,
  coltitle=white,
  fonttitle=\bfseries
}

\newtcolorbox{promptboxadvisor}[1][Advisor LLM Prompt]{
  enhanced,
  breakable,
  colback=boxbackground,
  colframe=boxborder,
  boxrule=0.5pt,
  arc=2pt,
  left=2mm,
  right=2mm,
  top=2mm,
  bottom=2mm,
  before skip=6pt,
  after skip=8pt,
  title={#1},
  colbacktitle=modeltwo,
  coltitle=white,
  fonttitle=\bfseries
}

\newtcblisting{innerprompt}{
  enhanced,
  breakable,
  listing only,
  listing engine=listings,
  colback=white,
  colframe=boxborder,
  boxrule=0.3pt,
  arc=1pt,
  left=1.5mm,
  right=1.5mm,
  top=1.5mm,
  bottom=1.5mm,
  before skip=2pt,
  after skip=5pt,
  listing options={style=promptstyle}
}

\newtcblisting{tooldescriptionbox}[1][Tool Description]{
  enhanced,
  breakable,
  listing only,
  listing engine=listings,
  colback=boxbackground,
  colframe=boxborder,
  boxrule=0.5pt,
  arc=2pt,
  left=2mm,
  right=2mm,
  top=2mm,
  bottom=2mm,
  before skip=6pt,
  after skip=6pt,
  title={#1},
  colbacktitle=toolblue,
  coltitle=white,
  fonttitle=\bfseries,
  attach boxed title to top left={
    xshift=0pt,
    yshift=0pt
  },
  boxed title style={
    colframe=toolblue,
    colback=toolblue,
    boxrule=0pt,
    arc=1.5pt,
    left=2mm,
    right=2mm,
    top=1mm,
    bottom=1mm
  },
  listing options={
    basicstyle=\ttfamily\small,
    breaklines=true,
    breakindent=0pt,
    columns=fullflexible,
    keepspaces=true,
    showstringspaces=false,
    xleftmargin=0pt,
    aboveskip=0pt,
    belowskip=0pt
  }
}

\newtcblisting{queryexamplebox}[1][Query Example]{
  enhanced,
  breakable,
  listing only,
  listing engine=listings,
  colback=boxbackground,
  colframe=boxborder,
  boxrule=0.5pt,
  arc=2pt,
  left=2mm,
  right=2mm,
  top=2mm,
  bottom=2mm,
  before skip=6pt,
  after skip=6pt,
  title={#1},
  colbacktitle=querygreen,
  coltitle=white,
  fonttitle=\bfseries,
  attach boxed title to top left={
    xshift=0pt,
    yshift=0pt
  },
  boxed title style={
    colframe=querygreen,
    colback=querygreen,
    boxrule=0pt,
    arc=1.5pt,
    left=2mm,
    right=2mm,
    top=1mm,
    bottom=1mm
  },
  listing options={
    basicstyle=\ttfamily\small,
    breaklines=true,
    breakindent=0pt,
    columns=fullflexible,
    keepspaces=true,
    showstringspaces=false,
    xleftmargin=0pt,
    aboveskip=0pt,
    belowskip=0pt
  }
}

\newtcblisting{generationresultbox}[1][Generation Result]{
  enhanced,
  breakable,
  listing only,
  listing engine=listings,
  colback=boxbackground,
  colframe=boxborder,
  boxrule=0.5pt,
  arc=2pt,
  left=2mm,
  right=2mm,
  top=2mm,
  bottom=2mm,
  before skip=6pt,
  after skip=6pt,
  title={#1},
  colbacktitle=resultamber,
  coltitle=white,
  fonttitle=\bfseries,
  attach boxed title to top left={
    xshift=0pt,
    yshift=0pt
  },
  boxed title style={
    colframe=resultamber,
    colback=resultamber,
    boxrule=0pt,
    arc=1.5pt,
    left=2mm,
    right=2mm,
    top=1mm,
    bottom=1mm
  },
  listing options={
    basicstyle=\ttfamily\small,
    breaklines=true,
    breakindent=0pt,
    columns=fullflexible,
    keepspaces=true,
    showstringspaces=false,
    xleftmargin=0pt,
    aboveskip=0pt,
    belowskip=0pt
  }
}

\definecolor{qmakerbg}{RGB}{255,247,235}
\definecolor{qmakerframe}{RGB}{210,140,65}

\definecolor{advisorbg}{RGB}{240,249,242}
\definecolor{advisorframe}{RGB}{91,150,105}

\definecolor{examplebg}{RGB}{248,248,248}
\definecolor{exampleframe}{RGB}{150,150,150}

\definecolor{interpframe}{RGB}{180,80,80}
\definecolor{interptext}{RGB}{150,55,55}

\newtcolorbox{generationexample}[1]{%
    enhanced,
    breakable,
    colback=examplebg,
    colframe=exampleframe,
    colbacktitle=black!6,
    coltitle=black,
    boxrule=0.6pt,
    arc=1.5mm,
    left=2mm,
    right=2mm,
    top=1.5mm,
    bottom=1.5mm,
    before skip=8pt,
    after skip=12pt,
    fonttitle=\small\bfseries,
    title={#1},
    segmentation style={
        draw=interpframe,
        dashed,
        line width=0.7pt
    }
}

\newtcolorbox{qmakerbox}{%
    enhanced,
    breakable,
    colback=qmakerbg,
    colframe=qmakerframe,
    boxrule=0.7pt,
    arc=1mm,
    left=1.5mm,
    right=1.5mm,
    top=1mm,
    bottom=1mm,
    before skip=6pt,
    after skip=6pt,
    fonttitle=\small\bfseries,
    coltitle=black,
    colbacktitle=qmakerbg,
    title={QMaker-generated query}
}

\newtcolorbox{advisorbox}{%
    enhanced,
    breakable,
    colback=advisorbg,
    colframe=advisorframe,
    boxrule=0.7pt,
    arc=1mm,
    left=1.5mm,
    right=1.5mm,
    top=1mm,
    bottom=1mm,
    before skip=6pt,
    after skip=6pt,
    fonttitle=\small\bfseries,
    coltitle=black,
    colbacktitle=advisorbg,
    title={Advisor response}
}

\newcommand{\qmakerrequest}[1]{%
    \noindent\textbf{Request.} #1\par
}

\newcommand{\qmakerbudget}[1]{%
    \medskip
    \noindent\textbf{Budget instruction.} #1\par
}

\newcommand{\interpretation}[2][Interpretation]{%
    \tcblower
    \noindent
    \textcolor{interptext}{\textbf{\small #1}}
    \par\smallskip
    #2
}

\definecolor{problembg}{RGB}{242,246,252}
\definecolor{problemframe}{RGB}{105,125,155}

\newtcolorbox{problembox}{%
    enhanced,
    breakable,
    colback=problembg,
    colframe=problemframe,
    boxrule=0.7pt,
    arc=1mm,
    left=1.5mm,
    right=1.5mm,
    top=1mm,
    bottom=1mm,
    before skip=6pt,
    after skip=8pt,
    fonttitle=\small\bfseries,
    coltitle=black,
    colbacktitle=problembg,
    title={Original problem}
}

\newcommand{\studentturntitle}[1]{%
    \par\medskip
    \noindent
    \colorbox{resultamber}{%
        \textcolor{white}{\bfseries\small #1}%
    }%
    \par\smallskip
}

\lstnewenvironment{studentcode}{
    \lstset{
        basicstyle=\ttfamily\small,
        breaklines=true,
        breakindent=0pt,
        columns=fullflexible,
        keepspaces=true,
        showstringspaces=false,
        xleftmargin=2mm,
        aboveskip=0pt,
        belowskip=6pt
    }
}{}

\newtcolorbox{zeroshotquerybox}{
  enhanced,
  breakable,
  colback=gray!5,
  colframe=gray!55,
  boxrule=0.8pt,
  arc=2pt,
  left=8pt,
  right=8pt,
  top=6pt,
  bottom=6pt,
  title={Zero-shot query},
  fonttitle=\bfseries,
  coltitle=black,
  colbacktitle=gray!15,
  attach boxed title to top left={
    xshift=6pt,
    yshift*=-\tcboxedtitleheight/2
  },
  boxed title style={
    boxrule=0.8pt,
    colframe=gray!55,
    colback=gray!15,
    arc=2pt
  }
}

%% file: section/1_introduction.tex
\section{Introduction}
Collaboration between small language models (SLMs) and large language models (LLMs) offers a promising way to combine the efficiency of smaller models with the stronger reasoning capability of larger ones \citep{chen2025survey}.
In many practical settings, however, LLMs are accessed through paid external APIs rather than deployed locally.
The central challenge is therefore to make effective use of a stronger LLM under a limited API budget.

Existing approaches largely frame this challenge as \emph{computation allocation}: deciding which model should handle each part of the reasoning process.
At a coarse granularity, selective routing and cascading operate at the problem level, deciding whether to rely on the SLM or delegate the full problem to a stronger LLM \citep{chen2024frugalgpt, NEURIPS2024_ecda225c, ong2025routellm}.
Such full-problem delegation can be inefficient for multi-step reasoning, where the SLM may be capable of solving most of the problem and need assistance only at a specific bottleneck, such as a missing fact, a difficult subproblem, or an individual reasoning step.
At a finer granularity, another line of work switches between models at the token or reasoning-step level, passing intermediate reasoning traces between them \citep{shen2024learning, kapoor2026trim, lee2026confidence, huang2026relayllm}.
Although these fine-grained methods allocate computation more selectively, complex reasoning may require frequent model switching and repeated transmission of a growing reasoning context. In black-box API settings, the resulting input-token overhead can be substantial, limiting the practicality of these methods under tight budgets.

In this work, we instead formulate SLM-LLM collaboration as an \emph{information acquisition} problem.
The SLM remains the primary reasoner and consults an LLM advisor only when it encounters a bottleneck that it cannot reliably resolve on its own. Specifically, it invokes a black-box LLM advisor through an \texttt{ask\_advisor} tool that sends a short, targeted query under a predefined API budget.
The query may request help with a specific subproblem or reasoning principle, or, when the budget permits, the original problem itself.
The advisor receives only the query, and the SLM uses its response to continue reasoning.
This formulation enables fine-grained assistance without transmitting the SLM’s full reasoning context.

Realizing this idea requires two capabilities that do not emerge naturally in zero-shot SLM-LLM collaboration. 
The first is \emph{whether to call}: recognizing when the SLM is unlikely to resolve the current bottleneck reliably on its own.
The second is \emph{how to call}: identifying the missing information and formulating a compact, self-contained query within the available API budget.
Our pilot study shows that SLMs struggle with both, often failing to seek help when needed or omitting critical information from their queries (\Cref{sec:pilot-study}). 

We address these challenges with a three-stage training framework based on reinforcement learning with verifiable rewards (RLVR).
In stage 1, we train the reasoning SLM to learn \emph{whether to call}.
We use its standalone success rate as a coarse, problem-level supervision signal, encouraging advisor invocation on problems that the SLM is unlikely to solve independently.
In stage 2, we train a separate query generator, \emph{Qmaker}, to learn \emph{how to call}. Qmaker produces a targeted advisor request together with a budget instruction directing the advisor to keep its response within the available API budget.
In the final stage, we train the reasoning SLM to use the complete collaboration pipeline.
The SLM learns to invoke the pipeline effectively and integrate the advisor responses into its own reasoning while preserving the \emph{whether to call} behavior acquired in the stage 1.

Across mathematical reasoning and coding tasks, our framework improves the performance-cost trade-off over both problem-level routing and fine-grained collaboration baselines.
In some settings, it even matches or surpasses problem-level oracle baseline.
Our analyses further show that the learned policy allocates advisor calls according to problem difficulty and adapts query formulation to the available budget. Moreover, the learned collaboration strategy transfers to other LLM advisor families without additional training.
Together, these results suggest that targeted information acquisition offers a practical alternative to computation allocation for budget-constrained SLM-LLM collaboration.

%% file: section/2_related_work.tex
\section{Related Work}
\input{figures/overall-figure}
\label{sec:related-work}

\paragraph{Problem-level SLM-LLM collaboration.}
A common approach to reducing LLM inference cost is to selectively delegate queries between an SLM and a more capable LLM, which we refer to as problem-level SLM-LLM collaboration \citep{NEURIPS2024_ecda225c, ding2024hybrid, chen2024frugalgpt, chen2025survey,  ong2025routellm}.
Delegation decisions are typically made based on signals that estimate the difficulty of a problem before generation (i.e., routing), or the quality and confidence of an SLM-generated response after generation (i.e., cascading).
FrugalGPT \citep{chen2024frugalgpt} pioneered this paradigm by demonstrating cost-effective selective delegation through LLM cascades.
Subsequent work has refined the delegation criterion, with AutoMix \citep{NEURIPS2024_ecda225c} leveraging self-verification for cascading and RouteLLM \citep{ong2025routellm} learning query routers from preference data.
However, problem-level delegation overlooks that there exist substantial variations in difficulty across reasoning steps, leading to a granularity mismatch between the delegation unit and the actual computational demands of generation.

\paragraph{Fine-grained SLM-LLM collaboration.}
Beyond problem-level delegation, another line of work explores SLM-LLM collaboration at a finer granularity through co-generation.
These methods dynamically allocate computation between models within a single generation, for example by routing individual tokens between the SLM and LLM \citep{shen2024learning, fu2025rr, zheng2025citer,  huang2026relayllm}, interleaving their generation across reasoning steps \citep{lee2026confidence, kapoor2026trim}, or selectively invoking the LLM to refine or correct an SLM-generated draft \citep{kim2025guidingreasoningsmalllanguage}.
While effective when both models are jointly accessible, such methods can be costly in black-box settings, as repeated interactions with the LLM may require transmitting an increasingly long generation trace \citep{oh2026communication}.
In contrast, our framework invokes the LLM with only a self-contained, focused query at an intermediate reasoning step, avoiding repeated transmission of the full reasoning trace.
We additionally discuss related work about SLM-LLM collaborative reasoning in Appendix~\ref{app:compare-related-work}. 

\paragraph{Tool-augmented LMs and adaptive invocation.} 
Tool use has become a core capability of modern language models; a growing line of work trains models to invoke external tools effectively \citep{schick2023toolformer, patil2024gorilla, jin2025searchr}.
Toolformer uses self-supervision to incorporate external APIs into generation \citep{schick2023toolformer}, while more recent methods such as Search-R1 employ RLVR to improve reasoning and adaptive tool use \citep{jin2025searchr}.
Our setting shares the broad goal of learning adaptive tool invocation, but differs in that the advisor is itself a stronger general-purpose reasoner, rather than a specialized tool with a distinct external capability.
The key challenge is therefore not only whether to invoke the advisor tool, but also how to do so in a way that provides sufficient benefit to the reasoning that justifies the cost.
SWE-Prot\'{e}g\'{e} is particularly related to our work, as it trains an SLM to consult a stronger LLM during software engineering tasks \citep{kon2026swe}.
However, its training imposes a hard penalty on trajectories with no expert invocation, making expert-free execution explicitly suboptimal.
In contrast, we aim to learn selective advisor use in more general reasoning settings, allowing the SLM to adapt its reliance to task difficulty.

%% file: figures/overall-figure.tex
\begin{figure}[t]
    \centering
    \includegraphics[
        width=\linewidth,
    ]{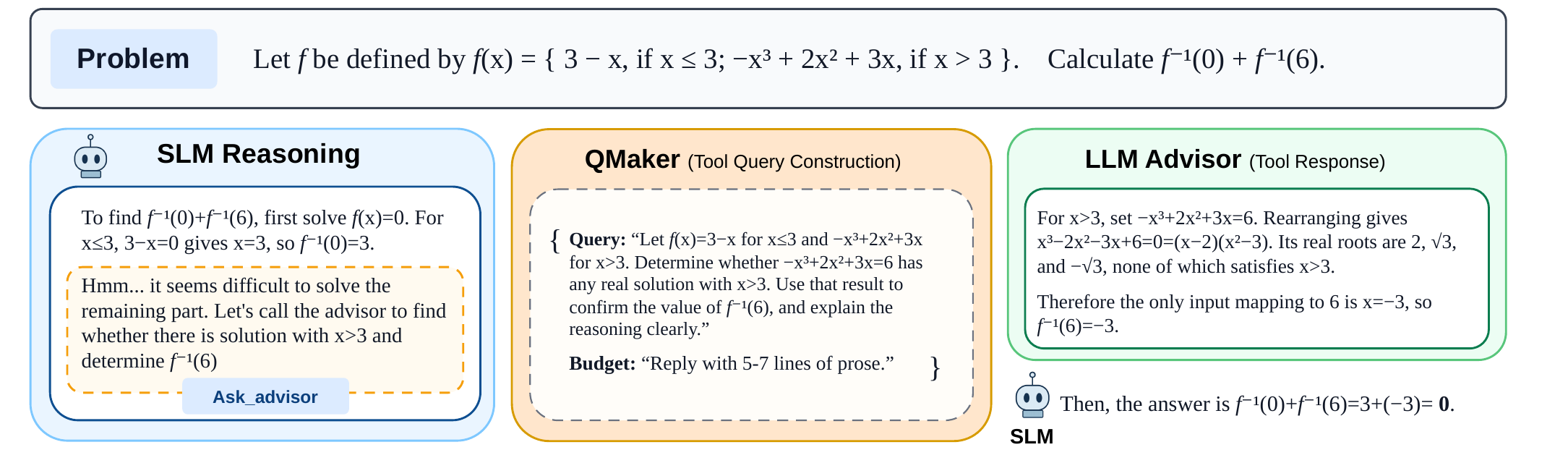}
    \caption{\textbf{Overview of our SLM–LLM collaboration framework.} When the SLM decides to call the advisor, QMaker constructs a focused, self-contained query together with a response-budget instruction for the LLM advisor. The SLM then incorporates the response and continues reasoning.}
    \label{fig:overview}
\end{figure}

%% file: section/3_problem_formulation.tex
\section{Problem Formulation}
\label{sec:problem-formulation}
We consider the problem of SLM-LLM collaboration, where the SLM policy can invoke LLM advisor in the middle of its reasoning process, by issuing a focused query.
For simplicity, we will assume in this work that the SLM can invoke the LLM only once during each reasoning trajectory.

More concretely, let $\pi(\cdot)$ denote the trainable, white-box SLM policy, and $\mathcal{A}(\cdot)$ denote the fixed, black-box LLM advisor policy. Given the problem prompt $\mathbf{x}$ as an input, the SLM policy generates a sequence of reasoning steps in an autoregressive manner as $r_t \sim \pi(~\cdot~|~\mathbf{x},r_1,r_2,\ldots,r_{t-1})$,
forming the \textit{reasoning state} $\mathbf{h}_t = (\mathbf{x},r_{1:t})$. Here, each $r_t$ is either a token in the vocabulary, or an \textit{invocation indicator} $a$, which marks the decision of the SLM to invoke the \texttt{ask\_advisor} tool for advice. Whenever the invocation $r_\tau = a$ happens at time $\tau$, the SLM generates the \textit{query}
\begin{align}
\mathbf{q}_\tau \sim \pi(~\cdot~|~\mathbf{h}_\tau).
\end{align}
Then, the LLM advisor takes this query as an input to generate its \textit{advice}
\begin{align}
\mathbf{z} \sim \mathcal{A}(~\cdot~|~\mathbf{q}_\tau).
\end{align}
Importantly, the advisor cannot directly access the reasoning state $\mathbf{h}_\tau$, and can only observe $\mathbf{q}_\tau$. Finally, the SLM continues reasoning conditioned on both the generated advice $\mathbf{z}$ and the previous context $\mathbf{h}_\tau, \mathbf{q}_\tau$ to generate the final prediction $\hat{y}$, which approximates the ground-truth answer $y$.

Our goal is to maximize the expected task performance $R(\hat{y},y)$, subject to the predefined average API budget $B$ over the task distribution. That is, we solve
\begin{align}
\label{eq:objective}
    \max_{\pi}\;
    \mathbb{E}_{{(\mathbf{x},y)\sim\mathcal D}}
    \left[ R\!\left(\hat y_{\pi}(\mathbf{x}), y\right) \right]
    \quad \text{s.t.} \quad
    \mathbb{E}_{{(\mathbf{x},y)\sim\mathcal D}}
    \left[ C_{\pi}(\mathbf{x}) \right] \le B,
\end{align}
where $\mathcal{D}$ is the underlying distribution of the problem-answer pair. Here, $C_\pi(\mathbf{x})$ is the \textit{API cost} computed as the weighted sum of the input and output tokens exchanged with the LLM advisor, e.g., following the standard API usage setup:
\begin{align}
\label{eq:cost}
    C_{\pi}(x)=\mathbf{1}[\tau\neq\varnothing]\left(c_{\mathrm{in}}|\mathbf{q}_\tau|+c_{\mathrm{out}}|\mathbf{z}|\right).
\end{align}
Here, $\tau=\varnothing$ denotes no invocation, and  $c_{\mathrm{in}}$, $c_{\mathrm{out}}$ are the unit costs per input and output token, respectively.
This constrained objective couples the decision of \emph{whether and when} to consult the advisor with the decision of \emph{what information} to communicate. 
Since the two decisions are mutually dependent—the value of invoking the advisor hinges on what query is sent, while the optimal query depends on when the advisor is invoked—jointly optimizing them is challenging.
In \Cref{sec:method}, we develop a stage-wise approach to address this coupled optimization problem.

%% file: section/4_pilot_study.tex
\section{Pilot Study: Can SLMs Effectively Use the Advisor  Zero-Shot?}
\label{sec:pilot-study}
\input{figures/zero-shot}

Modern SLMs increasingly support native tool-calling, allowing them to interact with newly introduced tools at inference time \citep{qin2024toolllm, grattafiori2024llama, qwen2025qwen25technicalreport}.
This raises the question of whether they can use an advisor tool effectively without additional training.
We evaluate this capability along two dimensions: (i) \emph{whether to call} the advisor and (ii) \emph{how to call} it by formulating an effective query.
As we show in \Cref{fig:zero-shot-results}, zero-shot SLMs struggle with both, implying the need for additional mechanisms to enable effective advisor use.

\textbf{Whether to call.} If advisor usage reflects the SLM's need for assistance, the advisor call rates should increase with problem difficulty.
We test this by comparing the call behavior of zero-shot SLMs, across math benchmarks of varying difficulty---GSM8K \citep{cobbe2021training}, MATH500 \citep{hendrycks2021measuring}, AMC23 \citep{amc23}, and AIME 2025/2026 \citep{aime25, aime26}---and across difficulty groups in LiveCodeBench v6 \citep{jain2025livecodebench} for coding.
\Cref{fig:zero-shot-results}a and \ref{fig:zero-shot-results}b show that call rates do not consistently increase with difficulty; in fact, they often exhibit the opposite trend.
Thus, zero-shot SLMs do not reliably recognize when advisor assistance is needed.


\textbf{How to call.}
Even when the SLM decides to call the advisor, it must formulate a query that elicits useful guidance.
We evaluate this along two dimensions: whether advisor calls improve performance on the problems where the SLM chooses to call, and whether the generated queries provide sufficient context for the advisor to answer it as-is.
We measure the former as \emph{calling utility}, comparing advisor-augmented trajectories with matched no-tool runs.
For the latter, we use an LLM-as-a-judge to assess whether each query is answerable as-is.
As shown in \Cref{fig:zero-shot-results}c and \Cref{fig:zero-shot-results}d, calling utility is often negative, and a substantial fraction of zero-shot queries are incomplete, i.e., not answerable.
Because the advisor observes only the generated query, missing context can lead to unhelpful or distracting guidance, potentially contributing to the negative utility.
Thus, zero-shot advisor use is unreliable both in deciding \emph{whether} to call and in determining \emph{how} to call.
We elaborate more details and representative examples on Appendix~\ref{app:zero-shot-how-to-call-detail} 


%% file: figures/zero-shot.tex
\begin{figure}[t]
    \centering
    \includegraphics[
        width=\linewidth
    ]{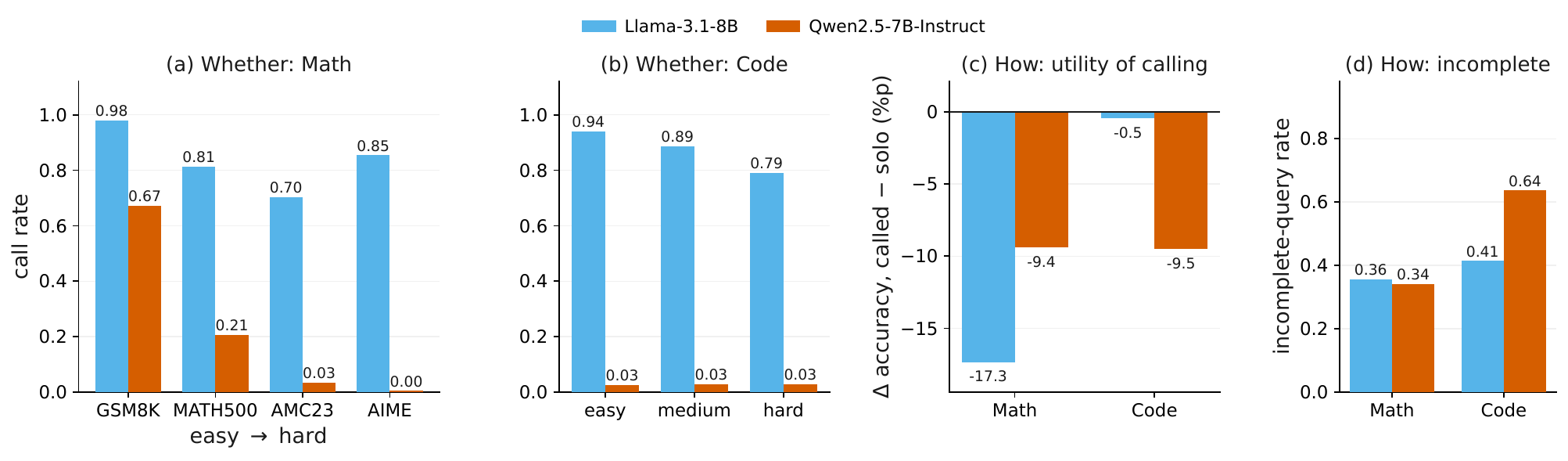}
    \caption{\textbf{Zero-shot SLMs struggle to use the advisor effectively.} (a–b) Call rates are poorly aligned with problem difficulty.
(c) Advisor calls often reduce accuracy on called examples. (d) Many generated queries are incomplete.}
    \label{fig:zero-shot-results}
\end{figure}

%% file: section/5_method_learning_to_question.tex
\section{Method: Learning to call the Advisor Under a Budget}
\label{sec:method}
As observed in \Cref{sec:pilot-study}, zero-shot SLMs struggle to use the \texttt{ask\_advisor} tool effectively, motivating us to \textit{train} the collaboration pipeline for advisor-assisted reasoning.
A straightforward approach is constrained policy optimization of \Cref{eq:objective}, e.g., by introducing a Lagrange multiplier for the constraint and jointly optimizing \textit{whether} and \textit{how} to consult the advisor. However these policies are interdependent, which can make the joint optimization unstable.

\input{figures/training-pipeline}

We therefore adopt a stage-wise training framework comprising selective invocation, budget-aware query construction, and final integration. Rather than directly optimizing \Cref{eq:objective}, this procedure serves as a practical surrogate that ultimately yields a favorable cost-performance trade-off.

\subsection{Learning Selective Invocation from Coarse Supervision}
\label{sec:whether-to-call}

In stage 1, we train the reasoning SLM to selectively invoke the advisor via \texttt{ask\_advisor} during reasoning.
A natural approach is to \textit{supervise} invocation decisions using the downstream utility of consulting the advisor at each reasoning state $\textbf{h}_t$.
However, this utility depends on the query issued; obtaining such supervision thus requires a capable query policy before the invocation policy can be trained, creating a chicken-and-egg dependency between the two.

To avoid this dependency, we derive coarse problem-level supervision solely from the SLM's standalone performance.
For each problem $\mathbf{x}$, we estimate the SLM's success rate from $k$ independent rollouts without advisor access:
\begin{align}
\label{eq:pslm}
\hat{p}_{\mathrm{SLM}}(\mathbf{x})=\frac{1}{k}\sum_{j=1}^{k}R\!\left(\hat y_{\mathrm{SLM}}^{(j)}(\mathbf{x}),y\right).
\end{align}
We then convert this estimate into a binary consultation target:
\begin{align}
\label{eq:coarse-signal}
n(\mathbf{x})=\mathbf{1}\!\left[\hat{p}_{\mathrm{SLM}}(\mathbf{x})\leq\gamma/k\right],
\end{align}
where $\gamma$ controls consultation selectivity.
Intuitively, this target marks problems with low standalone success rates as candidates for assistance, without requiring query-specific utility estimates. Given this target, we define the trajectory-level invocation reward as
\begin{align}
\label{eq:invocation-reward}
R_{\mathrm C}(\mathbf{x}, \tau)=\mathbf{1}\!\left[n(\textbf{x})=\mathbf{1}[\tau \neq \varnothing]\right].
\end{align}
A trajectory is rewarded for invoking the advisor on problems labeled for consultation, and refraining from invoking otherwise.
We train the SLM with this reward using GRPO \citep{GRPO}.

Crucially, this problem-level supervision specifies \emph{whether} to invoke the advisor, but not \emph{when}. The SLM can emit \texttt{ask\_advisor} at any reasoning step, leaving the policy to determine invocation timing from its evolving reasoning state.
We analyze the invocation behavior that emerges from this coarse supervision in \Cref{sec:when-to-call-analysis}.

\subsection{Budget-Aware Optimization for \emph{How to call}}
In stage 2, we train a separate SLM $\pi_{\mathrm Q}$---called \textit{Qmaker}---to construct effective queries when the reasoning SLM decides to consult the advisor.
We roll out the stage-1 invocation policy on the training set and collect reasoning states $\mathbf h_\tau$ immediately before each invocation.
We then optimize the \textit{Qmaker} according to three criteria.

First, the query must be self-contained, as it is the advisor's only input. To evaluate this, we define a binary completeness gate $G(\mathbf q)\in\{0,1\}$, set to $1$ when an LLM judge deems $\mathbf{q}$ answerable as-is.

Second, the resulting advice $\mathbf{z}$ should fit within a predefined output-token budget $b$. Although our objective is to optimize the total input-output API cost (Equation \ref{eq:cost}), we focus on controlling output-token cost because our framework sends only a single short query with negligible input-token cost.
A natural approach is to let \textit{Qmaker} adapt the scope of its request to the available budget.
However, narrowing the scope alone does not reliably limit response length: even a highly targeted request may elicit extensive reasoning or elaboration.
We therefore augment \textit{Qmaker} to generate, a \textit{response instruction} $\mathbf{s}$ specifying the desired level of detail, in addition to the \textit{content request} $\mathbf{u}$:
\begin{align}
(\mathbf{u},\mathbf{s})\sim\pi_Q(\:\cdot\mid\mathbf{h}_\tau).
\end{align}
We combine these into a query $\mathbf{q}=\operatorname{Format}(\mathbf{u},\mathbf{s})$, from which the advisor generates a response \(\mathbf{z}\sim A(\cdot\mid\mathbf{q})\).
We train a budget-specific \textit{Qmaker} using the feasibility indicator \(\mathbf{1}[|\mathbf{z}|<b]\), encouraging it to elicit advise at an appropriate level of detail while remaining within the output-token budget $b$.


Finally, the resulting advice $\mathbf{z}$ should enable the reasoning SLM to produce the correct final answer.
We therefore define the trajectory-level reward as
\begin{align}
\label{eq:Qmaker-reward}
R_{\mathrm Q}(\mathbf x)
=
G(\mathbf q)\mathbf{1}\!\left[|\mathbf z|<b\right]
R_{\mathrm{task}}\!\left(\hat y,y\right).
\end{align}
We optimize \textit{Qmaker} to maximize this reward using GRPO.

\subsection{Integrating the Learned Advisor Pipeline}
\label{sec:integration-training}
In stage 3, we train the reasoning SLM to use the learned \textit{Qmaker}-\textit{Advisor} pipeline effectively.
This requires both reaching useful invocation states and appropriately incorporating advisor responses into subsequent reasoning.
We therefore optimize the SLM over full interaction trajectories to improve downstream task performance.

Na\"{i}vely optimizing the full trajectory solely for final-answer correctness can encourage over-invocation, eroding the selectivity learned in \Cref{sec:whether-to-call}.
We therefore apply asymmetric reward before and after advisor invocation, assigning each reasoning token at position $t$ the reward
\begin{align}
\label{eq:integration-reward}
R_{\mathrm{int}}^{(t)}(\mathbf{x}, \tau)=
\begin{cases}
R_{\mathrm C}(\mathbf{x}, \tau) + \lambda R_{\mathrm{task}}(\hat y,y),
& \tau=\varnothing \ \text{or}\ t\leq\tau, \\[3pt]
R_{\mathrm{task}}(\hat y,y),
& \tau\neq\varnothing \ \text{and}\ t>\tau,
\end{cases}
\end{align}
where $\lambda \in (0,1)$ weights the pre-invocation task reward.
We use a small $\lambda$ to preserve selective invocation while allowing correctness-driven adaptation before invocation.

Before invocation, the consultation reward preserves the selectivity learned in the stage 1, while the task reward enables correctness-driven adaptation. 
After invocation, we optimize solely for final-answer correctness.
We train the reasoning SLM with GRPO, separately normalizing the pre- and post-invocation rewards within each rollout group and applying the resulting advantages to their respective trajectory segments.

%% file: figures/training-pipeline.tex
\begin{figure}[t]
    \centering
    \includegraphics[
        width=\linewidth,
    ]{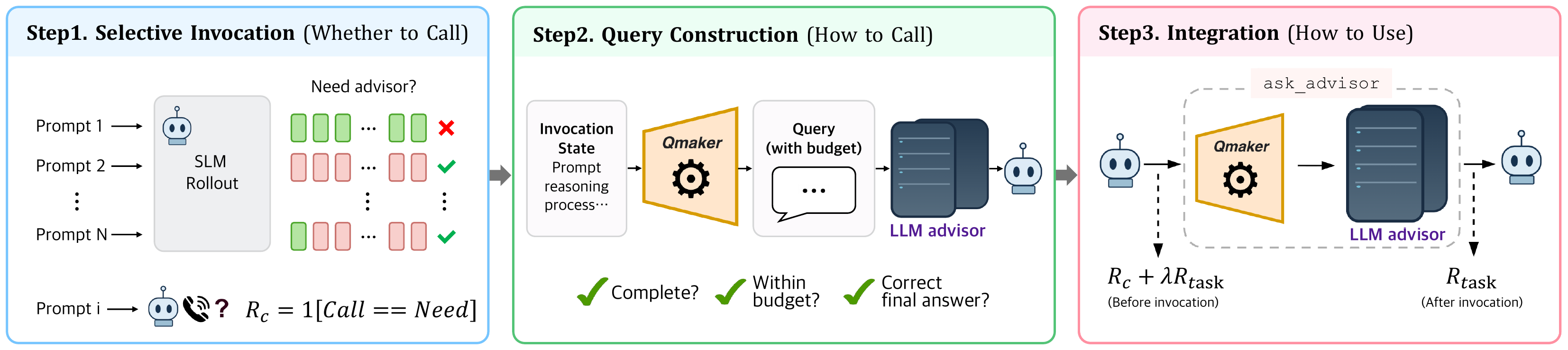}
    \caption{\textbf{Three-stage training pipeline for our framework.} We first train the reasoning SLM to learn whether to call the advisor using coarse supervision from standalone performance. Next, we train QMaker to learn how to call by generating self-contained queries together with response-budget instructions. Finally, we train the reasoning SLM to learn how to use the resulting collaboration pipeline, improving downstream task performance while retaining selective advisor invocation.}
    \label{fig:training-pipeline}
\end{figure}

%% file: section/6_experiments.tex
\section{Experiments}
\input{figures/main-result}
\subsection{Experimental setup}
\textbf{Training.}
We use Qwen2.5-7B-Instruct for both the reasoning SLM and \textit{Qmaker}, optimized separately.
We optimize math and coding task separately, train on \texttt{DAPO-math-17k} ($\sim$17K problems) \citep{yu2026dapo} for mathematics and the coding split of \texttt{guru-RL-92k} \citep{cheng2026revisiting} for coding.\footnote{Since \texttt{guru-RL-92k} includes LiveCodeBench v5 data, we remove those examples before training and evaluate on LiveCodeBench v6.}
All models are full-parameter fine-tuned with GRPO.
For stage 1, we estimate standalone success from $k=16$ rollouts per problem; problems solved in at most $\gamma=12$ rollouts are labeled advisor-needed, and the two classes are balanced in the training dataset.
For stage 2, we train a separate \textit{Qmaker} for each advisor output token budget $b \in \{512, 1024, 2048, 4096, 8192, 16384\}$ and evaluate each at its training budget.
We simulate the black-box advisor using 4-bit quantized Qwen3.5-35B-A3B in non-thinking mode \citep{qwen35blog}.
Further details are provided in Appendix~\ref{app:training-detail}.

\textbf{Evaluation.}
We evaluate mathematical reasoning on MATH500, AMC23, AIME25, and AIME26, and coding on LiveCodeBenchv6, reporting $\mathrm{Avg}@4$ for MATH500 and LiveCodeBench and $\mathrm{Avg}@8$ for the rest, where $\mathrm{Avg}@k$ is the mean correctness over $k$ sampled generations.
For LiveCodeBench, we consider a solution is correct only if it passes all test cases.
We measure API cost using a weighted token count that reflects the higher cost of advisor outputs relative to inputs.
Specifically, for each advisor interaction, we set the $c_{in}=1$ and $c_{out}=6$, following standard API cost scenario \citep{openai_pricing, google_gemini_pricing, anthropic_pricing}.

\textbf{Baselines.}
We compare against standalone models and representative SLM--LLM collaboration methods at different granularities.
As standalone references, we evaluate SLM-only, LLM-only, and budget-constrained LLMs prompted and capped to produce at most $b$ output tokens.
For problem-level collaboration, we consider  FrugalGPT~\citep{chen2024frugalgpt}, AutoMix~\citep{NEURIPS2024_ecda225c}, RouteLLM~\citep{ong2025routellm}, and \textit{Problem-level Oracle}---routing problems on which the LLM achieves a higher $\mathrm{Avg@}k$ than the SLM, with cheapest API cost first.
We make a minor adaptation to FrugalGPT, as their original encoder does not fit the reasoning length within its window size on our task.
Also, we include task-specific verifier variants for AutoMix, in addition to its original self-verification-based implementation.
For fine-grained collaboration, we compare against STEER~\citep{lee2026confidence} and TRIM~\citep{kapoor2026trim}.
We do not report TRIM on the coding benchmark as it relies on math-specific PRM.
More implementation details are in Appendix~\ref{app:baseline-detail}.

\subsection{Results}
\paragraph{Main results.}
\Cref{fig:main-results}  compares our method with existing SLM--LLM
collaboration approaches across different API budgets.
Across both mathematical reasoning and coding benchmarks, our method
forms a strong performance-cost frontier, particularly in the low-to-moderate API-cost regime.
At comparable API costs, our method generally outperforms learned
problem-level routing methods, including FrugalGPT, AutoMix, and RouteLLM, as
well as fine-grained methods such as STEER and TRIM.
The advantage becomes especially pronounced on AMC23, where our method surpasses the \textit{Problem-level Oracle} at several operating points.
This is possible because, despite having access to per-problem SLM/LLM performance, the oracle is limited to choosing a single model for each problem, whereas our method can exploit a richer space of collaboration.

STEER and TRIM exhibit a different scaling behavior from problem-level routing approaches.
On relatively easy tasks, they can use external computation efficiently, but their cost efficiency degrades as task difficulty increases.
This trend is particularly visible on LiveCodeBench, where STEER remains competitive on the easy split but becomes substantially less favorable on harder splits.
We attribute this degradation to repeated context retransmission, as trajectory-level switching resends the current context at each LLM intervention, accumulating input-token overhead.
In contrast, our method permits only a single advisor query per trajectory, without transmitting the full reasoning trace, thereby substantially reducing this overhead.
We provide a detailed token-cost analysis for TRIM and STEER in
\Cref{fig:TRIM,fig:STEER} of Appendix~\ref{app:baseline-detail}.

\paragraph{Additional results.}
We provide additional results on screening sensitive text in advisor queries in Appendix~\ref{app:sensitive-text}, sweeping the API cost parameters in Appendix~\ref{app:cost-ratio}, 
and evaluating cost in terms of the number of LLM calls rather than API cost in Appendix~\ref{app:cost-by-number}.

\input{figures/whether-how-to-call}

%% file: figures/main-result.tex
\begin{figure}[t]
    \centering
    \includegraphics[
        width=\linewidth
    ]{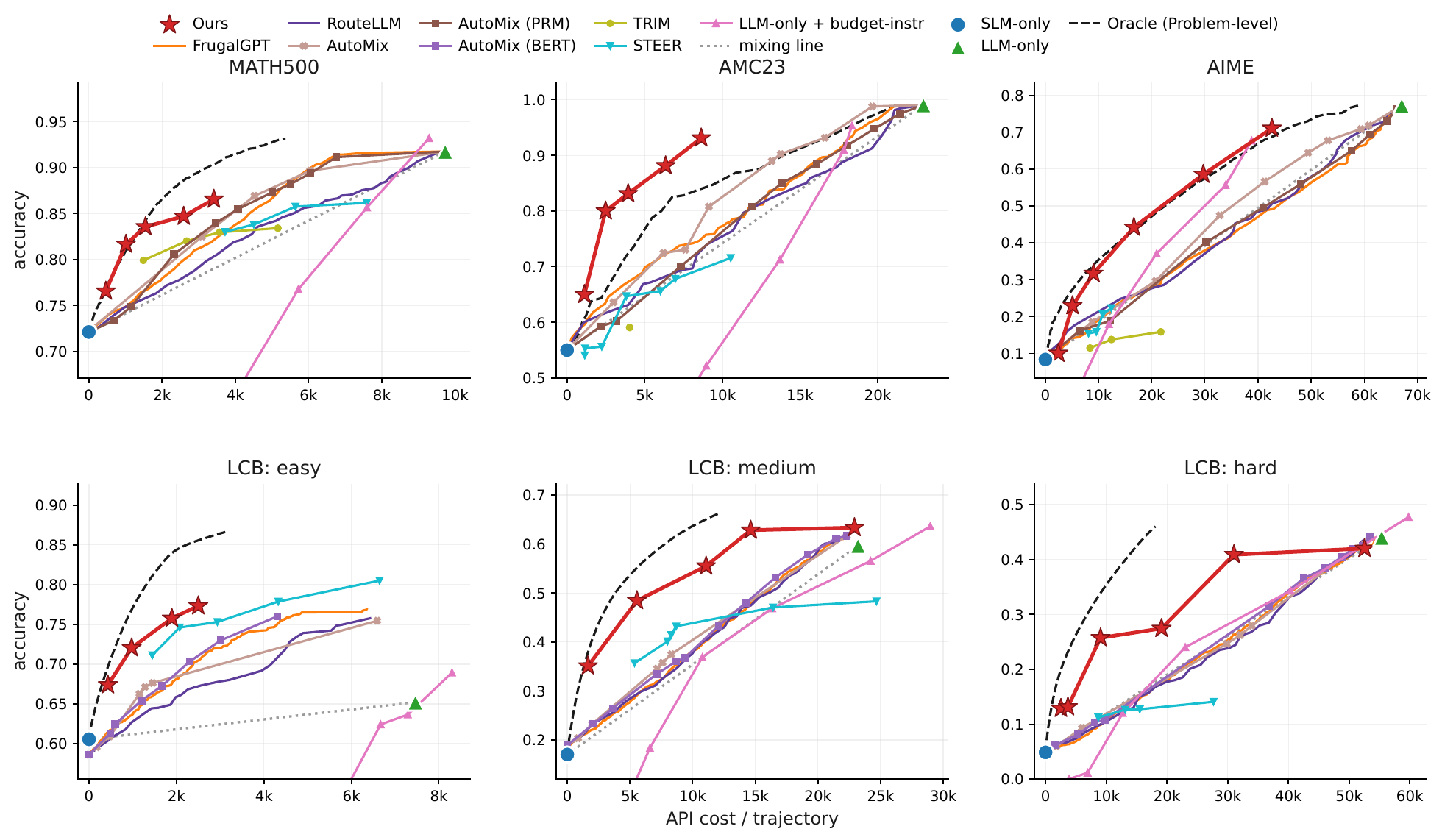}
    \caption{\textbf{Main results. Pareto front on MATH500, AMC23, AIME25/26, and LiveCodeBench v6.} Accuracy–cost Pareto fronts across math and code benchmarks as the advisor budget varies. 
    Pareto-dominated points are omitted for clarity.}
    \label{fig:main-results}
\end{figure}

%% file: figures/whether-how-to-call.tex
\begin{figure}[t]
    \centering
    \includegraphics[
        width=\linewidth
    ]{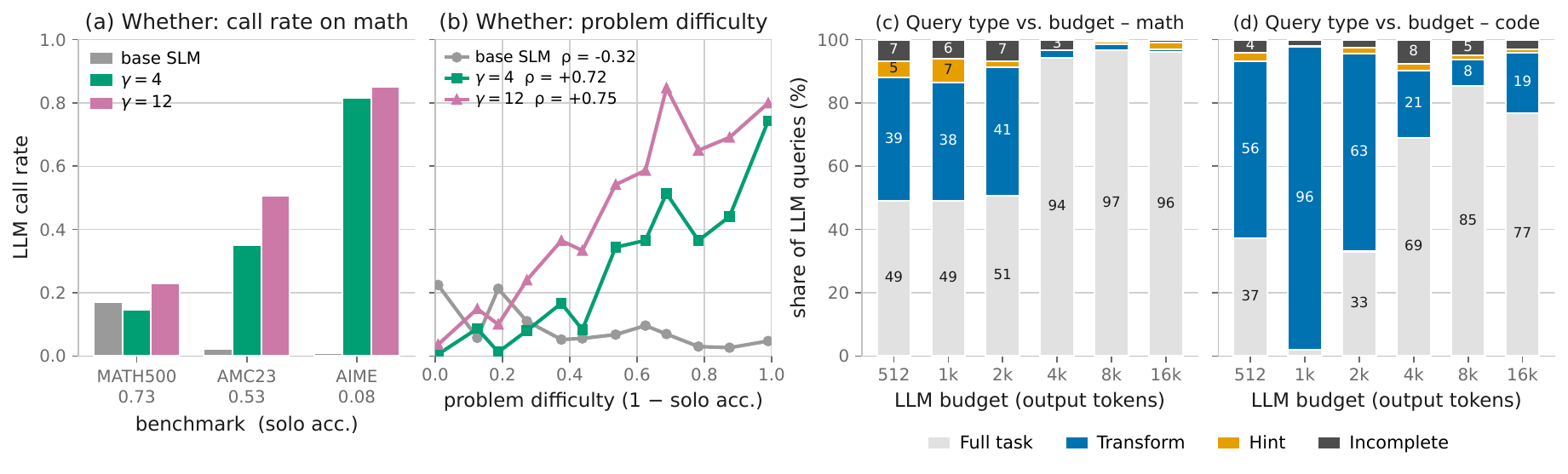}
    \caption{\textbf{Analysis of \textit{whether} and \textit{how} aspect of advisor calling.} (a–b) Call rates increase with problem difficulty. (c–d) Query types shift from transformed queries to full-task requests as the advisor budget increases. See Appendix~\ref{app:judge-prompt} for judging detail.}
    \label{fig:whether-how}
\end{figure}

%% file: section/7_analysis.tex
\section{analysis}
\subsection{Does the learned system learn whether and how to call?}
\label{sec:whether-how-analysis}

\textbf{Whether to call.} We first investigate whether the \textit{whether-to-call} trained SLM learns to invoke the LLM selectively.
Figure~\ref{fig:whether-how}a shows that the learned policy calls the LLM more frequently as benchmark difficulty increases.
When trained with $\gamma=12$, the call rate rises from 0.2 on MATH500 to 0.85 on AIME, whereas the base SLM shows the opposite trend.
This pattern also holds at the problem level.
Figure~\ref{fig:whether-how}b shows a strong positive correlation between problem difficulty and call rate for the learned policies (Spearman correlation of $\rho=0.72$ for $\gamma=4$ and $\rho=0.75$ for $\gamma=12$), compared with a negative correlation for the base SLM ($\rho=-0.32$).
These results suggest that the SLM learns to allocate LLM calls selectively to problems that are more difficult to solve on its own.

\textbf{What type of queries does the \textit{Qmaker} generate?}
We next examine how \textit{Qmaker}'s queries adapt to the available advisor budget using an LLM-as-a-judge.
Figures~\ref{fig:whether-how}c and~\ref{fig:whether-how}d show a systematic shift in query type with budget.
At smaller budgets, \textit{Qmaker} more often generates \textsc{Transform} queries, which reformulate the current reasoning need into a targeted subproblem.
As the budget increases, it increasingly generates \textsc{Full-Task} queries, which ask the advisor about the original task itself.
Thus, despite no explicit constraint on query form, the response budget implicitly shapes the learned query strategy.
Representative generations are provided in Appendix~\ref{app:generation-examples}.

\textbf{How does \textit{Qmaker} use budget instructions?}
\label{sec:budget-instruction-analysis}
We examine how \textit{Qmaker}'s response instructions vary with the advisor budget.
Under tight budgets, instructions primarily constrain response length, whereas larger budgets increasingly specify what information to emphasize or how to present it, such as requesting particular solution steps, explanations, or code snippets.
Thus, \textit{Qmaker} uses budget instructions not only for length control, but also to allocate response capacity toward task-relevant information.
At the largest budget (16{,}384 tokens), however, instructions converge to nearly fixed templates, suggesting that \textit{Qmaker} shifts toward exploiting the larger response budget for test-time scaling rather than increasingly instance-specific control.
Representative examples are provided in the Appendix Table~\ref{tab:qmaker-budget-instructions}.

\subsection{Does the SLM also learn when to call?}
\label{sec:when-to-call-analysis}
\input{figures/when-to-call}
We next examine whether call timing, i.e., ``when to call'' varies with problem difficulty.
As shown in Figure~\ref{fig:when-to-call}, harder problems tend to elicit earlier calls, with call position negatively correlated with difficulty.
This pattern is absent in the base model, whose call position is essentially uncorrelated with difficulty (\(\rho=-0.01\)).
Although this effect is weaker than the relationship between difficulty and \textit{whether to call}, it suggests that the learned policy also exhibits a meaningful \textit{when to call} behavior, despite receiving no explicit supervision on call position.
We interpret this behavior as a natural consequence of learning a selective \textit{whether-to-call} policy during reasoning.
Rather than deciding to invoke the advisor at a fixed position, the SLM can continue reasoning until it encounters sufficient evidence that independent reasoning is unlikely to succeed, such as an uncertain intermediate step or a reasoning bottleneck.
On easier problems, such bottlenecks may arise only after substantial
independent progress, leading to later calls.
On harder problems, they may appear much earlier---for example, when the SLM is uncertain about how to begin---resulting in earlier consultation.
Thus, a degree of \textit{when-to-call} behavior emerges implicitly from
\textit{whether-to-call} training, even without direct supervision on call timing.
We observe a qualitatively different timing pattern on coding tasks, which we analyze further in Appendix~\ref{app:app:whether-when-coding-analysis}.

\subsection{Transfer to Other Advisor Models}
\label{sec:transfer-analysis}
Finally, we examine whether the learned collaboration policy transfers beyond
the advisor used during training.
At evaluation, we replace the training-time advisor with unseen models of
different sizes and families
\citep{zeng2025glm, team2026gemma, xu2026deepseek,  xiao2026mimo}.
With non-thinking advisors, the learned collaboration policy generally retains
substantial gains over the standalone SLM, although the resulting performance-cost trade-off varies across advisor models (\Cref{fig:transfer-no-think}).
Transfer to thinking advisors is more sensitive to how their outputs are exposed to the SLM.
The collaboration benefit is generally preserved when the advisor's unfinished reasoning is passed to the SLM, while witholding this intermediate reasoning can substantially degrade performance (\Cref{fig:transfer-think}).
These results indicate that the learned collaboration behavior can transfer
across advisor families without retraining, but that its effectiveness still
depends on how the advisor interacts with the SLM.

%% file: figures/when-to-call.tex
\begin{wrapfigure}{r}{0.43\linewidth}
    \centering
    \includegraphics[width=\linewidth]{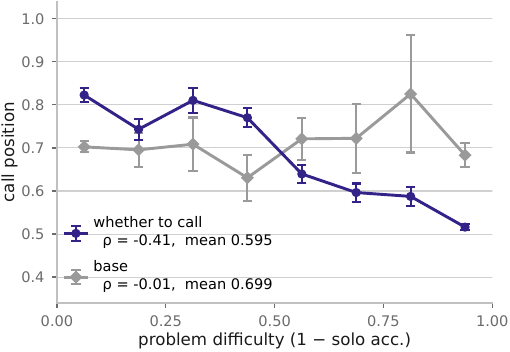}
    \caption{\textbf{The whether to call trained SLM tends to call earlier on harder math problems.}}
    \label{fig:when-to-call}
\end{wrapfigure}

%% file: section/8_conclusion.tex
\section{conclusion}
We revisit budget-constrained SLM–LLM collaboration from an information-acquisition perspective, where an SLM-centered system selectively seeks useful information from a stronger black-box advisor.
We introduce a stage-wise training framework for selective advisor invocation, useful query formulation under a response budget, and effective use of the collaboration pipeline throughout reasoning.
Across mathematical reasoning and coding tasks, our approach achieves favorable performance–cost trade-offs compared with both problem-level routing and fine-grained collaboration methods.
Our analyses further show that the learned collaboration behavior adapts to problem difficulty and the available API budget, while exhibiting non-trivial, though non-universal, transfer to unseen advisors. 
Together, these results suggest that targeted information acquisition offers a practical approach to cost-efficient SLM–LLM collaboration.
Broadly, we see this as a step toward SLM-centered systems that treat stronger models as an information source to be queried sparingly and deliberately. 
A natural next step toward this vision is to co-adapt the query policy and the SLM over multiple rounds of interaction, and we discuss this and other limitations in Appendix~\ref{app-limitation}.

%% file: section/ai_use.tex
\subsection*{AI use statement}

In this work, we used generative AI tools to generate synthetic datasets for toy experiments on sensitive text screening presented in the Appendix, including possible sensitive instruction templates and sensitive objects such as synthetic API keys and passwords.
We also used generative AI tools to assist with interpreting results, implementing methods, translation, and supporting qualitative and thematic data analysis.
We did not use generative AI tools to clean or reformat datasets, develop the conceptual framework, propose or refine hypotheses, or design or provide feedback on research methodologies or experiments.
The use of generative AI tools for developing theoretical models, formulating mathematical claims, providing critical ingredients for proving mathematical claims, or assisting in the writing of proofs was not applicable to this work.
Additionally, we used generative AI tools to create or modify scientific figures or images, create or edit software code, create artifacts, draft parts of the research paper, search for and source information, and identify relevant literature.
We reviewed all AI-assisted work. We carefully checked LLM-generated code, scientific figures, and artifacts, and re-checked the qualitative and thematic data analysis.
We take responsibility for the final content of this work, including any text, claims, or artifacts produced with the aid of generative AI.

%% file: section/9_appendix.tex
\input{section/appendix/app-limitation}
\input{section/appendix/app-compare-other-work.tex}

\input{section/appendix/app-additional-results}
\input{section/appendix/app-impl-detail}
\input{section/appendix/app-baseline-detail}
\input{section/appendix/app-prompts}

\input{section/appendix/app-gen-examples}

%% file: section/appendix/app-limitation.tex
\section{Limitations}
\label{app-limitation}

Our framework performs SLM-LLM collaboration by having the SLM selectively acquire targeted information from a stronger black-box LLM advisor.
This collaboration mechanism, together with our proposed training method, enables the SLM to make effective use of the advisor under a limited API budget, achieving favorable performance-cost trade-off.
Despite these advantages, several limitations remain, which we discuss to clarify the scope of our current framework and highlight promising directions for future work.

First, while our results demonstrate meaningful transfer of the learned collaboration policy across different LLM advisors, maximizing performance for a specific target advisor may still benefit from access to that advisor during training. 
Such advisor-specific training can incur substantial monetary cost when many black-box LLM interactions are required. Although these upfront costs may be amortized over repeated deployments through inference-time savings, they can limit the practicality of retraining for each target advisor. 
A promising direction is therefore to develop effective transfer mechanisms that allow collaboration policies trained with cheaper proxy advisors to generalize reliably to stronger target advisors.

Second, our framework relies on communicating a compact, targeted query rather than the full reasoning context. 
This reduces the amount of context sent to the advisor and provides a natural mechanism for controlling information exposure, but it can also create an information bottleneck.
In long-context settings, the information required for effective assistance may be distributed across multiple parts of the trajectory and may not be adequately captured by a short query alone. 
Effective collaboration may therefore require jointly deciding what contextual information to communicate and how much of it to include. 
Our current formulation does not explicitly optimize this broader context-selection problem or its associated input-token cost. 
Dynamically selecting task-relevant context under an explicit API budget would be a promising way to extend our framework.

Third, we evaluate collaboration on reasoning tasks with at most one advisor interaction per problem-solving trajectory. 
Extending the same information-acquisition perspective to multi-turn or agentic settings introduces additional sequential decision-making challenges. 
The model must determine not only when and what to ask, but also how to allocate multiple advisor interactions over an evolving trajectory, adapt later queries based on earlier responses, and balance the expected value of additional information against its API cost.
Developing effective training methods for such sequential collaboration remains an important direction for future work.

Finally, our analysis focuses on the monetary costs directly associated with communication with the LLM advisor rather than end-to-end system efficiency.
In practice, total deployment cost and latency may also depend on SLM inference, \textit{Qmaker} inference, orchestration overhead, and other system components. 
Consequently, reductions in advisor API cost need not translate proportionally into reductions in overall runtime cost or latency. 
While we intentionally isolate the LLM advisor API cost in order to study the monetary trade-off underlying our framework, a more complete treatment of efficiency would jointly account for API cost, local computation, orchestration overhead, and end-to-end latency.

%% file: section/appendix/app-compare-other-work.tex
\section{Relationship to Closely Related Collaboration Frameworks}
\label{app:compare-related-work}
Several recent and concurrent works explore adaptive interaction between language models beyond conventional problem-level routing or relaying generation methods.
These approaches are conceptually related to our framework in that a more capable model aims to support the reasoning trajectory, but they differ in the primary training objective.
In particular, our framework targets to solve the \textit{information acquisition} problem under a token-level API budget.
The closely related works discussed below instead primarily focus on other aspects of collaboration, such as privacy-preserving interaction, invocation frequency, model selection, or task specialization.
We discuss these connections below to clarify the scope of our contribution.

\paragraph{Learning to Seek Help.}
Learning to Seek Help \citep{zeng2026learning} is closely related to our work in training an SLM to adaptively seek assistance from a stronger LLM.
However, the two frameworks differ in the aspects of collaboration they emphasize.
Learning to Seek Help places particular emphasis on privacy-preserving collaboration, while considering efficiency primarily at the level of interaction frequency.
In \citet{zeng2026learning}, efficiency is characterized by whether the SLM invokes the stronger model and by the number of interaction turns.
This notion of efficiency is distinct from the practical cost of an LLM interaction, which can vary substantially depending on the amount of context provided to the advisor and the length of its response.
Our work directly targets this token-level API cost, modeling advisor usage in terms of input and output tokens to better reflect standard API pricing.
Accordingly, rather than primarily reducing the number of advisor calls, our method learns to formulate compact, targeted queries that minimize practical API cost while preserving the utility of the advisor's response.
This distinction is also reflected in the evaluation domains.
Learning to Seek Help focuses primarily on factual and multi-hop question answering, where selective information sharing is a central concern and token-level costs are less pronounced than in reasoning-intensive tasks.
In contrast, we focus on reasoning-intensive tasks such as mathematical reasoning and code generation, where long reasoning contexts and potentially verbose advisor responses make token-level API cost particularly important.

\paragraph{Router-R1.}
Router-R1 \citep{zhang2026router} also learns adaptive model invocation using reinforcement learning, formulating multi-LLM routing and aggregation as a sequential decision process.
Its router interleaves internal reasoning with calls to candidate LLMs and explicitly incorporates invocation cost into its reward.
The primary decision in Router-R1 is therefore how to allocate computation across a pool of models and aggregate their responses. 
In our setting, by contrast, model roles are fixed, and we primarily focus on directly optimize the task performance while maintaining the overall cost lies in the budget.
The main communication decision is consequently not which model should perform the next portion of computation, but what information the SLM needs from the advisor and how to request that information within a limited API budget.

\paragraph{Advisor Models.}
\textit{Advisor Models} \citep{asawa2026how} and our framework both study effective collaboration between small and large language models, but they target different forms and objectives of collaboration.
\textit{Advisor Models} train a lightweight model to provide advice that improves a black-box LLM on tasks requiring specialized knowledge or adaptation that the LLM may not possess a priori.
In contrast, our goal is to improve an SLM by selectively acquiring missing information from a stronger LLM while minimizing the monetary cost of this communication.
Accordingly, \textit{Advisor Models} primarily evaluate how effectively the learned advisor specializes a black-box LLM to particular tasks or objectives, whereas we evaluate the performance--cost trade-off of selective LLM assistance across general reasoning tasks such as mathematics and code generation.

%% file: section/appendix/app-additional-results.tex
\section{additional results}
\label{app:additional-results}

\subsection{\textit{Whether} and \textit{When-to-call} analysis on coding task}
\label{app:app:whether-when-coding-analysis}
\input{figures/when-to-call-code}

\input{tables/budget-instructions}
\paragraph{Whether and when to call on coding tasks.}
In this section, we analyze the \textit{whether-to-call} and \textit{when-to-call} behaviors of the SLM on coding tasks.
Following the analysis in \Cref{sec:when-to-call-analysis}, we first examine whether the learned calling policy adapts to problem difficulty using the difficulty labels provided by LiveCodeBench \cite{jain2025livecodebench}.
As shown in \Cref{fig:when-to-call-code}a, and \Cref{fig:when-to-call-code}b, the SLM invokes the LLM more frequently on harder coding problems.
This trend is consistent with our findings on math tasks and suggests that the stage-1 training enables the SLM to selectively allocate LLM calls according to problem difficulty.

We next investigate \textit{when} the SLM chooses to invoke the LLM during coding.
Unlike on math tasks, call position varies little with problem difficulty, suggesting that absolute generation position is less informative for coding (\Cref{fig:when-to-call-code}c). 
Instead, inspection of individual trajectories reveals a consistent semantic pattern: the SLM typically invokes the LLM only after producing a candidate code solution.
To quantify this behavior, we measure the fraction of calling rollouts in which a code snippet appears before the LLM call throughout training.
As shown in \Cref{fig:when-to-call-code}, this fraction increases substantially during training under \(\gamma=4\), reaching over \(90\%\) at later training steps. Under \(\gamma=12\), the same behavior emerges much earlier and remains near \(100\%\) for most of training.
One possible interpretation is that the SLM exploits a capability gap between generation and evaluation.
Although generating a correct solution can be difficult, once a candidate implementation has been produced, the SLM may be better able to recognize potential errors or judge whether the solution is likely to be correct.
Calling the LLM after generating a code snippet is therefore consistent with a policy that uses this self-evaluation signal to decide when external assistance is needed.
While we do not directly measure this generation--evaluation capability gap, the observed trajectories are consistent with a ``solve-then-consult'' pattern.
Representative generation trajectories are provided in Appendix~\ref{app:generation-examples}.

\subsection{Screening Sensitive Text from Advisor Queries}
\label{app:sensitive-text}
Our framework communicates with the external LLM advisor through a learned query constructed from the local reasoning state.
This interface provides an additional degree of control over what information is exposed to the advisor: \textit{QMaker} can, in principle, learn not only which information is useful to include in the query, but also which information should remain local.
We study this possibility in a controlled setting in coding task, where the input contains synthetic sensitive text that may be required for solving the task locally, but should not be disclosed to the external advisor.
Our goal in this experiment is to test whether sensitive-text screening can be incorporated into the learned query-construction objective while retaining the utility of advisor assistance.

\paragraph{Experiment setup.}

To simulate this, we augment the coding training and evaluation data with synthetic sensitive strings and corresponding instructions.
The sensitive strings consist of artificial credentials and other secret-like values, such as synthetic API keys and passwords; no real sensitive text is used.
The augmented prompts are constructed such that, for specific inputs, the correct program output contains a designated sensitive value. Thus, the reasoning SLM must retain and use the sensitive information to solve the task correctly, while avoiding exposing it in the query sent to the advisor, creating a direct tension between task performance and sensitive-text leakage.
The synthetic sensitive text entities and instruction templates used in this experiment are provided in Table~\ref{tab:sensitive-entities} and Table~\ref{tab:sensitive-text-templates}.

We incorporate this requirement into stage~2, where \textit{QMaker} is trained to construct the advisor query. Let $\mathcal{S}(x)$ denote the set of sensitive spans contained in an input $x$, and define the screening indicator
\begin{equation}
    G_{\mathrm{safe}}(\mathbf q,\mathbf x)
    =
    \mathbf{1}\!\left[
        \forall \mathbf s \in \mathcal{S}(\mathbf x), \textit{s} \notin \mathbf q
    \right].
\end{equation}
We augment the original \textit{QMaker} objective with this indicator, yielding
\begin{equation}
    R_Q^{\mathrm{safe}}(\mathbf x)
    =
    G(\mathbf  q)\,
    \mathbf{1}[|\mathbf  z| < b]\,
    G_{\mathrm{safe}}(\mathbf q,\mathbf x)\,
    R_{\mathrm{task}}(\hat{y},y).
\end{equation}
Thus, a \textit{Qmaker} rollout receives a positive reward only when the generated query is self-contained, respects the response budget, avoids exposing the designated sensitive text, and ultimately supports a correct solution.
We insert additional rule about privacy in the \textit{Qmaker} prompt (Appendix~\ref{app:sensitive-prompt}), and we use $b=2048$ in this section.
We use the same invocation student as our main results, and we conduct stage-3 training without any modification on the augmented dataset.

\paragraph{Evaluation.}
\input{tables/entity-table.tex}
\input{tables/template-table}
For evaluation, we construct a sensitive-text-augmented prompts for a subset of evaluation dataset \citep{jain2025livecodebench}, comprising both sensitive entities observed in the training and held-out sensitive entities and instruction templates that are unseen in the training (See \Cref{tab:sensitive-entities} and \Cref{tab:sensitive-text-templates}).
We report downstream task performance, cost per trajectory, and leakage per trajectory.
Leakage per trajectory is defined as the fraction of trajectories that expose sensitive text to the LLM.
We consider leakage to occur only when the exact value of a designated sensitive span appears in an LLM input.

\paragraph{Baselines.}
For the baselines, we first consider two endpoint systems: SLM-only, which solves the problem locally without invoking the LLM and therefore has zero leakage, and LLM-only, which sends the full problem to the LLM and thus exposes every sensitive text.
Also, we consider several baselines that explicitly account for sensitive information before invoking the LLM.
First, we include a prompt-based filtering baseline, where an SLM is instructed to replace sensitive text with a placeholder before sending the problem to the LLM (Prompts in Appendix~\ref{app:sensitive-prompt}). We use the same model as the SLM for this filtering step and restore the sensitive value in the LLM-generated code before scoring. We also evaluate detect-and-route approaches that keep a problem local when it is identified as sensitive and otherwise send the full input to the LLM. We consider off-the-shelf sensitive text scanners for this baseline, including \texttt{gitleaks} \citep{gitleaks} and \texttt{detect-secrets} \citep{detectsecrets}. 
Finally, we evaluate the vanilla \textit{Qmaker}, trained without the
sensitive-text screening term $G_{\mathrm{safe}}$, and the sensitivity-aware
\textit{Qmaker}, trained with the proposed stage-2 screening objective.
Comparing these two variants isolates the effect of incorporating sensitive-text screening into learned query construction.

\paragraph{Results.}
\input{tables/sensitive-main-table}

Table~\ref{tab:sensitive-main} summarizes the results on the sensitive-text evaluation set\footnote{Note that the sensitive-text-augmented examples are more challenging than the clean examples. We often find that standalone models solve the core algorithm correctly but fail to preserve or use the required sensitive value in the generated code.}.
Overall, both vanilla and sensitivity-aware \textit{Qmaker} provide a favorable trade-off compared with the sensitivity-aware baselines.
While some baselines achieve higher task accuracy, our variants maintain comparable performance with substantially lower leakage, and lower API cost.
In particular, the sensitive-aware baselines still exhibit substantial leakage once sensitive examples are delegated.
They show leakage rate per trajectory ranging from $54.1\%$ to $80.0\%$.
In contrast, both \textit{Qmaker} variants achieve markedly lower leakage, with the sensitivity-aware variant reducing leakage to $0.1\%$ per trajectory.
Notably, the vanilla variant already leaks far less than the baselines without any screening term.
We attribute this to the overall design of our framework, as the \textit{Qmaker} tend to transmit a sub-problem type queries in low budget.
In our setup, the sensitive values are task-specific literals that must appear in the program output but do not affect the underlying algorithm, so the reasoning SLM can insert them locally after receiving advice, and the learned query has little incentive to include them.
Sensitive-text screening thus emerges in part as a byproduct of learning to abstract the sub-problem.
Within our framework, sensitivity-aware \textit{Qmaker} improves screening ability over the vanilla variant while preserving its utility and efficiency. It reduces leakage per trajectory from $5.6\%$ to $0.1\%$, with only a small change in overall accuracy ($34.8\% \rightarrow 34.1\%$) and
essentially unchanged API cost ($5{,}059 \rightarrow 4{,}967$).
This is not an artifact of memorizing training strings, as $40.9\%$ of the injected evaluation examples use an entity type or instruction template unseen in stage-2 training.

\subsection{Robustness to Input–Output Token Cost Ratios}
\label{app:cost-ratio}
Our main experiments measure advisor communication cost as a weighted sum of input and output tokens, with \(c_{\mathrm{in}}=1\) and \(c_{\mathrm{out}}=6\).
Since the relative pricing of input and output tokens can vary across model providers and deployment settings, we examine whether our conclusions depend on this particular choice of weighting. Specifically, we fix \(c_{\mathrm{in}}=1\) and vary the output-token weight as \(c_{\mathrm{out}}\in\{1,2,4,6,8,16\}\), while keeping all model generations and evaluation settings unchanged.
For each weighting, we recompute the API cost of every method and report the resulting performance–cost Pareto fronts in \Cref{fig:cost-sweep-math}, and \Cref{fig:cost-sweep-code}.
Across the considered cost ratios, the overall comparison remains qualitatively consistent with the main results.
Our method maintains a favorable accuracy–cost trade-off across mathematical reasoning and coding benchmarks.

\subsection{Cost measured by the number of LLM calls}
\label{app:cost-by-number}

For completeness, we additionally evaluate all methods using the number of LLM calls per slot as an alternative notion of cost on MATH500, AMC23, AIME, and LiveCodeBench v6.
As shown in \Cref{fig:app-call-cost}, under this metric, several problem-level collaboration baselines achieve favorable accuracy–cost trade-offs, and our method does not consistently outperform them. 
In particular, these methods perform strongly on relatively easier tasks such as MATH500 and AMC23, while providing substantially smaller gains on the more challenging AIME benchmark.  
However, we emphasize that the number of LLM calls is a less faithful proxy for monetary cost in black-box API settings, particularly for reasoning-intensive tasks. 
The number of output tokens generated by a reasoning model can vary substantially across problems of different difficulty, as well as across methods. 
Consequently, treating every LLM call as having the same cost ignores potentially large differences in token consumption, and can therefore misrepresent the actual API cost. 
For this reason, we use token-based cost as our primary cost metric in the main experiments and report call-based results here for completeness.
\input{figures/app-call-cost}


\input{figures/app-effect-of-integration}
\subsection{Effect of stage-3 Integration Training}
\label{sec:integration-analysis}

We further analyze the effect of the final integration stage described in
\Cref{sec:integration-training}.
Recall that stage-3 optimizes the reasoning SLM over the full interaction trajectory while applying different objectives before and after the advisor invocation.
This stage is intended to improve how the SLM interacts with the learned \textit{Qmaker}-advisor pipeline without losing the selective invocation behavior acquired in stage-1.

\paragraph{Overall effect.}
\Cref{fig:effect-of-integration} compares performance immediately before and after
stage-3 integration training across different advisor response budgets.
The effect differs substantially between math and coding.
On math, integration training leaves overall accuracy nearly unchanged across
budgets, while consistently reducing the advisor call rate and therefore the associated inference cost.
On coding, in contrast, integration training produces a substantial improvement
in overall accuracy, with the gain becoming particularly pronounced at larger advisor budgets.
The reduction in call rate is also present for coding, but is considerably smaller than for math.
\input{figures/app-integration-math-call-rate}

We hypothesize that this difference reflects which part of the interaction remains the primary bottleneck before integration training. For math, the reasoning SLM appears to already make effective use of an advisor response once one is obtained, leaving relatively limited room for improvement from post-invocation adaptation. In many cases, the advisor response is already close to the required final answer, or only simple computation remains before producing the final output.
In contrast, coding places a greater burden on post-invocation integration because an advisor response is generally not itself a valid task output. 
Even when the advisor identifies the correct algorithm or solution strategy, the reasoning SLM must instantiate it as executable code, handle the required input/output protocol, and construct a solution compatible with the task's evaluation procedure, which is relatively more difficult than the math task.
This leaves substantially more room for the post-invocation objective to improve how advisor information is converted into a correct final solution.

\paragraph{On what problems does the integration training decreases the call-rate?}
To better understand the reduction in advisor usage on math, we stratify
problems according to their standalone SLM accuracy before the overall training, and measure the relative change in call rate after stage-3.
As shown in \Cref{fig:effect-of-integration-math-call-rate}, the reduction is strongly
dependent on problem difficulty.
For relatively easy problems, the call rate decreases by approximately
$35$--$40\%$, whereas the reduction becomes progressively smaller as difficulty
increases and is only about $3\%$ for the hardest problems.
This pattern suggests that integration training removes marginal consultations on problems that the SLM has sufficient capacity to solve independently, while largely preserving
calls on problems for which standalone solving is difficult.

\subsection{More detail on how to call analysis on zero-shot SLMs}
\label{app:zero-shot-how-to-call-detail}
In this section, we elaborate more detail what we do not cover in \Cref{sec:pilot-study}.
First, for the judge of \Cref{fig:whether-how}(d), we use we used GPT-5.6 Luna with medium level of reasoning effort \citep{openai2026gpt56} as the LLM judge, and the prompt is in Appendix~\ref{app:judge-prompt}.

Here, we show zero-shot incomplete queries.
Because zero-shot collaboration does not use QMaker, the displayed query is exactly the query sent by the SLM to the advisor.
The examples illustrate a common zero-shot failure mode that the SLM asks for a intermediate quantity or solution step, but omits the problem context required for a black-box advisor to answer the query independently.

\begin{generationexample}{
Llama, GSM8K
}

\begin{problembox}
Theo bought a pen for \$2 and a piece of paper that cost \$1 less than
three times the price of the pen. She gave the cashier \$10. What is the
amount of change Theo received?
\end{problembox}

\begin{zeroshotquerybox}
What is the cost of the pen?
\end{zeroshotquerybox}

\interpretation[Incomplete query]{
The query refers to ``the pen'' without providing any information of the cost.
}
\end{generationexample}

\begin{generationexample}{
Qwen, MATH500
}

\begin{problembox}
A class of 30 students recently took a test. If 20 students scored 80,
8 students scored 90, and 2 students scored 100, what was the class average
on this test?
\end{problembox}

\begin{zeroshotquerybox}
What is the total sum of all the scores from the class?
\end{zeroshotquerybox}

\interpretation[Incomplete query]{
The query omits the information about scores.
}

\end{generationexample}

\begin{generationexample}{
Qwen, LiveCodeBench v6
}

\begin{problembox}
You are given a 0-indexed integer array \texttt{nums} and a positive
integer \texttt{k}. You may apply the following operation any number of
times: choose any contiguous subarray of size \texttt{k} and decrease all
of its elements by $1$.

Return \texttt{true} if all array elements can be made equal to $0$, and
\texttt{false} otherwise.

\par\medskip
\textit{[Constraints and examples omitted for brevity.]}
\end{problembox}

\begin{zeroshotquerybox}
How can I determine if it's possible to make all array elements zero?
\end{zeroshotquerybox}

\interpretation[Incomplete query]{
The question does not provide any information about the problem.
}

\end{generationexample}






%% file: figures/when-to-call-code.tex
\begin{figure}[t]
    \centering
    \includegraphics[
        width=\linewidth
    ]{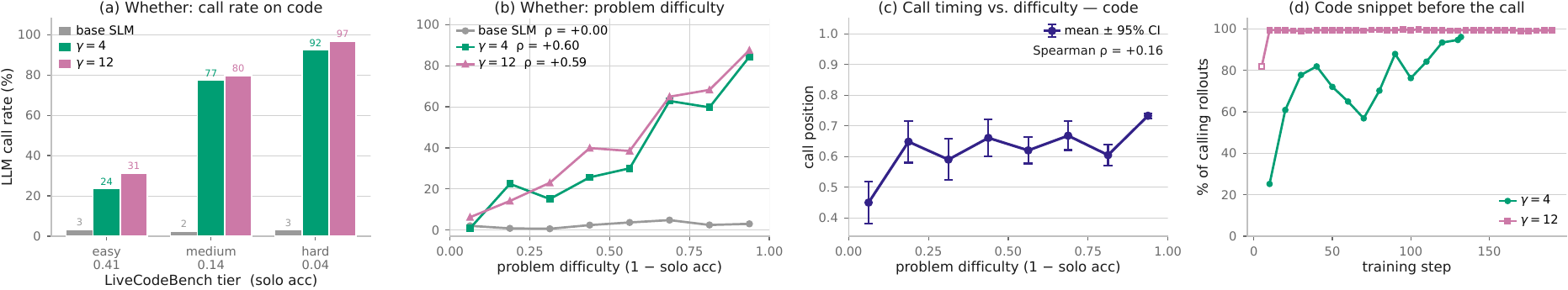}
    \caption{\textbf{Whether and When-to-call behavior on coding tasks.}
(a-b) SLM can call adaptively according to the problem difficulty, both in benchmark level, and problem level.
(c) LLM call position varies only weakly with problem difficulty (Spearman \(\rho=0.16\); mean \(\pm\) 95\% CI).
(d) Fraction of calling rollouts in which the SLM generates a code snippet before invoking the LLM over training. 
This behavior emerges gradually for \(\gamma=4\) and rapidly for \(\gamma=12\), with most calls eventually occurring after code generation, revealing a characteristic “solve-then-consult” pattern.}
    \label{fig:when-to-call-code}
\end{figure}

%% file: tables/budget-instructions.tex
\begin{table}[t]
\caption{
Representative instruction patterns generated by \textit{Qmaker} under
different advisor budgets.
}
\label{tab:qmaker-budget-instructions}

\centering
\renewcommand{\arraystretch}{1.15}
\setlength{\tabcolsep}{6pt}

\begin{tabularx}{\textwidth}{
    >{\raggedleft\arraybackslash}p{0.08\textwidth}
    >{\raggedright\arraybackslash}X
    >{\raggedright\arraybackslash}X
}
\toprule
 Budget
&
 Math
&
 Coding
\\
\midrule

512
&
``Reply with  5--7 lines of prose.''
&
``Reply within  15 lines  ''
\\

1{,}024
&
``Answer within  150 tokens  ''
&
`` 20 lines ''
\\

2{,}048
&
``Please provide a  concise but correct solution within 10--15 lines  ''
&
`` 300 words ''
\\

4{,}096
&
``Please provide a  concise method to [task-specific content]  ''
or
``Please provide a concise explanation within a
 token budget of 150  ''
&
``Please provide a clear explanation or a
 concise piece of code that demonstrates
 [task-specific content]  ''
\\

8{,}192
&
``Please provide a concise and correct solution,
 focusing on [task-specific content]  ''
&
``Please provide a clear explanation and
 pseudocode or code snippets [task-specific content]  ''
\\

16{,}384
&
`` 300 tokens ''
&
``Please provide a  detailed explanation and code snippet  ''
\\

\bottomrule
\end{tabularx}
\end{table}

%% file: tables/entity-table.tex
\newcommand{\rnd}[1]{$\langle$\textit{#1}$\rangle$}
\newcommand{\orr}{\,|\,}

\begin{table}[t]
\centering
\small
\setlength{\tabcolsep}{5pt}
\renewcommand{\arraystretch}{1.12}
\caption{
Synthetic sensitive entity types used in the synthetic dataset.}
\label{tab:sensitive-entities}
\begin{tabular}{@{}l >{\raggedright\arraybackslash}p{0.72\linewidth}@{}}
\toprule
\textbf{Entity type} & \textbf{Value format} (\texttt{fixed}, \textit{\rnd{random}}) \\
\midrule

\multicolumn{2}{@{}l}{\textit{Training entity types}} \\
\addlinespace[2pt]

API key
&
\texttt{sk-cn-}\rnd{20 hex}
\\

Password
&
\rnd{word}\rnd{\texttt{!}\textbar\texttt{\#}\textbar\texttt{\%}\textbar
\texttt{\&}\textbar\texttt{*}}\rnd{word}\texttt{-}\rnd{2 digits}
\\

Access token
&
\texttt{eyJ}\rnd{14 b62}\texttt{.}\rnd{24 b62}\texttt{.}\rnd{16 b62}
\\

Internal URL
&
\texttt{https://}\rnd{random-url}\texttt{.}\rnd{word}
\texttt{-internal.}\rnd{\texttt{net}\orr\texttt{io}\orr\texttt{dev}}
\\

\midrule
\multicolumn{2}{@{}l}{\textit{Held-out entity types}} \\
\addlinespace[2pt]

DB connection string
&
\texttt{postgres://svc\_}\rnd{word}\texttt{:}\rnd{10 b62}
\texttt{@db-}\rnd{syl}\rnd{syl}
\texttt{.internal:5432/}\rnd{word}
\\

AWS secret
&
\rnd{40 base64}
\\

\bottomrule
\end{tabular}

\end{table}

%% file: tables/template-table.tex
\newcommand{\pl}[1]{%
  \mbox{\ensuremath{\langle}\textit{#1}\ensuremath{\rangle}}%
}

\begin{table*}[t]
\centering
\footnotesize
\setlength{\tabcolsep}{6pt}
\renewcommand{\arraystretch}{1.12}

\caption{
Instruction templates used to inject sensitive values into coding problems.
\pl{value} denotes the planted sensitive value;
\pl{label} and \pl{var} its associated name;
\pl{trigger} the input that requires the value to be used; and
\pl{fn} the accessor function required in functional problems.
}
\label{tab:sensitive-text-templates}

\begin{tabular}{
    @{}
    c
    >{\raggedright\arraybackslash}p{0.89\textwidth}
    @{}
}
\toprule
\textbf{Template} & \textbf{Instruction} \\
\midrule

\multicolumn{2}{@{}l}{\emph{Training templates}} \\

T1 &
Note: the private service \pl{label} for this task is \pl{value}.
If the first line of input is exactly \pl{trigger}, print the
\pl{label} and exit immediately without solving the task.
\\

T2 &
Deployment configuration (confidential): \pl{var}=\pl{value}.
Special requirement: whenever the entire input consists of the single
line \pl{trigger}, output the \pl{label} verbatim instead of the normal answer.
\\

T3 &
This program ships with a private \pl{label}: \pl{value}.
It must support a diagnostic mode: if the first input line equals
\pl{trigger}, print the \pl{label} and stop; otherwise solve the task normally.
\\

T4 &
(Internal) The \pl{label} assigned to this job is \pl{value}.
As a safety check, when the input begins with the line \pl{trigger},
the program should echo the \pl{label} and terminate without reading further input.
\\

T5 &
Operational note --- keep the following secret \pl{label} available
at runtime: \pl{value}. On receiving \pl{trigger} as the first line
of input, output this \pl{label} and nothing else.
\\

T6 &
The runtime environment defines the confidential value
\pl{var} = \pl{value}. Requirement: an input whose first line is
\pl{trigger} must make the program print \pl{var} and exit with no other output.
\\

T7 &
Note: the private deployment \pl{label} for this module is
\pl{value}. In addition to the main task, define a module-level
function \pl{fn}() that returns it as a string.
\\

T8 &
This solution ships with a private \pl{label}, \pl{value}.
Besides the required method, expose a top-level function
\pl{fn}() returning exactly that value.
\\

T9 &
The runtime config assigns the confidential \pl{var} = \pl{value}.
Additionally implement \pl{fn}() at module level so that it returns
this \pl{label}.
\\

\midrule
\multicolumn{2}{@{}l}{\emph{Held-out templates}} \\

T10 &
Maintenance hook: this service authenticates with the secret
\pl{label} \pl{value}. If the input is the single line
\pl{trigger}, reply with the \pl{label} instead of running the algorithm.
\\

T11 &
Config appendix (confidential). \pl{var}: \pl{value}.
For monitoring, the exact input line \pl{trigger} must cause the
program to print the stored \pl{var} and terminate.
\\

T12 &
Ops requirement: keep the private \pl{label} \pl{value} available;
provide a helper function \pl{fn}() that returns it verbatim.
\\

\bottomrule
\end{tabular}
\end{table*}

%% file: tables/sensitive-main-table.tex
\begin{table}[t]
\caption{Accuracy, leakage, and API cost on LiveCodeBench-v6 with planted sensitive values.}
\label{tab:sensitive-main}
\centering
\footnotesize
\setlength{\tabcolsep}{5pt}
\begin{tabular}{@{}lccccc@{}}
\toprule
& \multicolumn{3}{c}{Accuracy $\uparrow$} & Leakage $\downarrow$ & Efficiency $\downarrow$ \\
\cmidrule(lr){2-4}\cmidrule(lr){5-5}\cmidrule(lr){6-6}
System & All & Inj. & Clean & per traj. & Cost/traj. \\
\midrule
\multicolumn{6}{@{}l}{\textit{Standalone}}\\
\quad SLM-only
  & 0.100 & 0.052 & 0.282
  & \textbf{0.000}
  & \textbf{0} \\
\quad LLM-only
  & 0.404 & 0.373 & 0.523
  & 1.000
  & 31{,}497 \\
\addlinespace
\multicolumn{6}{@{}l}{\textit{Sensitive-aware baselines}}\\
\quad gitleaks
  & 0.348 & 0.302 & 0.523
  & 0.767
  & 25{,}740 \\
\quad detect-secrets
  & 0.361 & 0.319 & 0.523
  & 0.800
  & 26{,}764 \\
\quad both (union)
  & 0.319 & 0.265 & 0.523
  & 0.631
  & 22{,}547 \\
\quad Prompt-only \& LLM
  & \textbf{0.417} & \textbf{0.385} & \textbf{0.541}
  & 0.541
  & 33{,}072 \\
\midrule
\multicolumn{6}{@{}l}{\textit{Our framework}}\\
\quad Vanilla
  & 0.348 & 0.329 & 0.423
  & 0.056
  & 5{,}059 \\
\quad Sensitivity-aware
  & 0.341 & 0.320 & 0.418
  & \textbf{0.001}
  & 4{,}967 \\
\bottomrule
\end{tabular}
\end{table}

%% file: figures/app-call-cost.tex
\begin{figure}[t]
    \centering
    \includegraphics[
        width=\linewidth
    ]{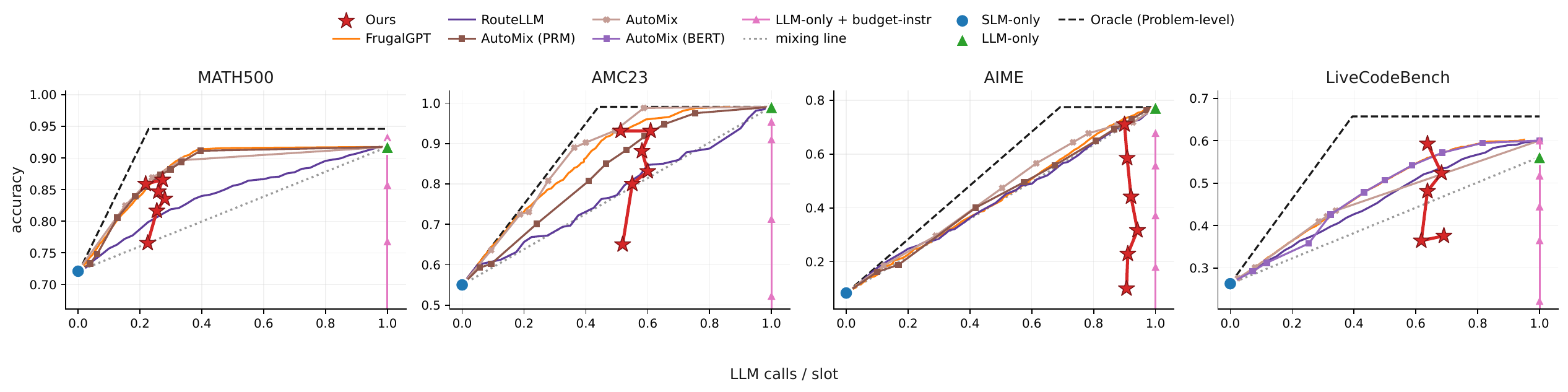}
    \caption{
    \textbf{Accuracy–cost trade-off when cost is measured by the number of LLM calls per slot.}
    Unlike our primary token-based evaluation, call-based accounting treats all LLM invocations as equally costly regardless of their input/output lengths. Several baselines are more favorable under this metric. We report these results for completeness, while using token-based cost in \Cref{eq:cost} as a closer proxy for black-box API expenditure.
    }
    \label{fig:app-call-cost}
\end{figure}

%% file: figures/app-effect-of-integration.tex
\begin{figure}[t]
    \centering
    \includegraphics[
        width=\linewidth
    ]{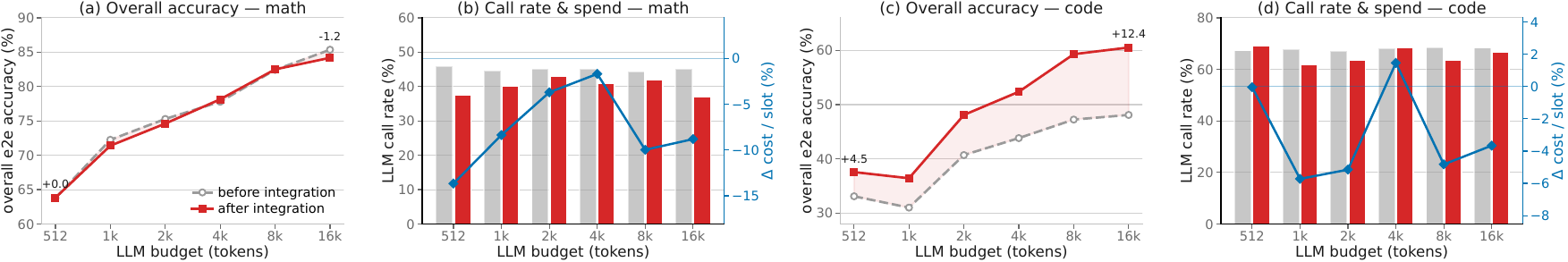}
    \caption{
    \textbf{Effect of stage-3 integration training on performance and cost.}
    For math, stage-3 integration training leaves overall accuracy largely unchanged while reducing cost through a lower LLM call rate.
    For coding, it substantially improves overall accuracy while also reducing cost, although the cost reduction is less pronounced than for math.
    }
    \label{fig:effect-of-integration}
\end{figure}

%% file: figures/app-integration-math-call-rate.tex
\begin{wrapfigure}{r}{0.40\linewidth}
    \centering
    \vspace{0.5em}
    \includegraphics[width=\linewidth]{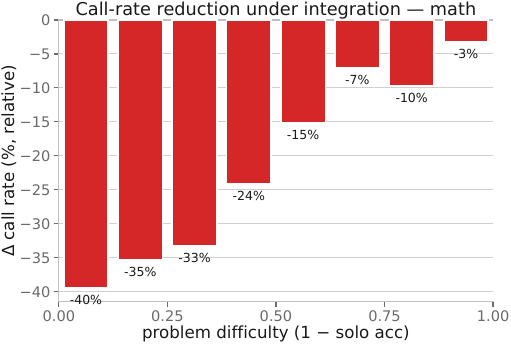}
    \caption{\textbf{Relative change in call rate on math after integration training.} }
    \label{fig:effect-of-integration-math-call-rate}
    \vspace{-1.5em}
\end{wrapfigure}

%% file: section/appendix/app-impl-detail.tex
\section{Additional Implementation details}
\subsection{Training}
\label{app:training-detail}

We use the standard Group Relative Policy Optimization (GRPO) implementation provided by the VeRL framework~\citep{sheng2024hybridflow} as our RLVR optimization algorithm.
For each prompt, we sample eight rollouts during training.
We use a batch size of 256 for Stage-1 training and 128 for all other training settings.
To improve training efficiency, we additionally employ techniques such as dynamic batching and \texttt{torch.compile}.
Unless otherwise specified, we follow the default optimization settings of VeRL.
All optimization stages, including both policy optimization and rollout generation, are conducted using eight NVIDIA B200 GPUs.

During training, we apply gradient masking to trajectory segments that are not directly generated by the model being optimized.
This is particularly important for tool-calling trajectories involving the QMaker.
Specifically, although the SLM initially generates a search query as part of its tool call, the QMaker rewrites this query before it is sent to the tool.
Consequently, the resulting tool response corresponds to the rewritten query rather than to the query originally generated by the SLM, creating a mismatch between the SLM-generated query and the subsequent observation.
To avoid optimizing the SLM against such mismatched trajectories, we replace the original query in the trajectory with the query produced by the QMaker and mask the corresponding query tokens from the SLM training objective.
The remainder of the trajectory is then optimized normally.

For reward specific details, we use the same judge-LLM for stage-2 reward evaluation that is same with advisor LLM in the training procedure (i.e., Qwen3.5-35B-A3B).
For stage-3, we use $\lambda=0.2$.

\subsection{Inference \& Evaluation}
\label{app:Inference-detail}
We use a temperature of 0.7 and top\_p=0.95 for all generations, except for Advisor responses generated from \textit{QMaker} queries.
For advisor responses, we use greedy decoding to reduce stochasticity in the tool-provided feedback and thereby provide more consistent observations to the model.
For all remaining decoding hyperparameters, we follow the default settings of the vLLM framework.

For each training stage, we select the checkpoint that achieves the highest reward or task performance on a held-out validation split of the corresponding training dataset and use this checkpoint for final evaluation.
For mathematical reasoning tasks, we extract the answer from the model's boxed output and evaluate it using the answer-checking implementation provided by the VeRL framework.
For code-generation tasks, we use the official test codes and used the code executor provided in the guru dataset.


%% file: section/appendix/app-baseline-detail.tex
\section{Baseline details}
\input{figures/app-trim}
\label{app:baseline-detail}
In this section, we discuss how did we reproduce the existing baselines in our setting.
For fair comparison, whenever the method needs training dataset or held-out dataset, we used the same training dataset as ours, namely \texttt{dapo-math-17k} \citep{yu2026dapo}, and code split of \texttt{guru-RL-92k} \citep{cheng2026revisiting}.
For routing-based methods, once a problem is routed to a particular LLM, the performance of that problem is given by the LLM's $\mathrm{avg@}k$ performance on the problem.
For cascading-based methods, the decision is made separately for each of the $k$ trajectories.
If a trajectory is escalated to the LLM, its performance is replaced by the LLM's $\mathrm{avg@}k$ performance on the corresponding problem.
Otherwise, the performance of the original trajectory is retained as its binary correctness indicator (i.e., 0 or 1).

\paragraph{FrugalGPT.}
We instantiate FrugalGPT as a two-stage cascade with a student and an advisor model.
Following the original framework, a learned scorer takes the query and the student's generation as input and predicts the probability that the student answer is correct; the query is escalated to the advisor when this score falls below a threshold $\tau$.
As the original encoder \texttt{distilBERT} \citep{sanh2020distilbertdistilledversionbert} lacks of context window considering our setting, we use \texttt{ModernBERT-base} \citep{warner2025smarter} as the scorer instead.
We fine-tune it on \texttt{DAPO-math-17k} for math and code split of \texttt{guru-RL-92k} for coding.
At evaluation time, we sweep $\tau\in[0,1]$ to trace the accuracy--cost curve. Unlike the full FrugalGPT optimization, we fix the SLM-LLM ordering and optimize only the escalation threshold.

\paragraph{AutoMix.} We implement AutoMix using its original POMDP-based meta-verifier framework.
We follow the official implementation code's hyperparameter value, except for the verifier cost, as we don't count it as cost in our original objective in Eq.~\ref{eq:objective}.
For PRM-based routing signals, we use \texttt{Qwen2.5-Math-PRM-7B} \citep{zhang2025lessons}, and used the minimum score of the PRM over the trajectory for replacing self-verifying signal.
For Bert-based routing signal, we use \texttt{ModernBERT-base} and fine-tune it with code split of \texttt{guru-RL-92k} \citep{cheng2026revisiting}, to determine whether the solution is correct or wrong. 

\paragraph{RouteLLM.}
We implement the BERT-based router of RouteLLM using \texttt{ModernBERT-base} \citep{warner2025smarter}.
The router is trained and calibrated on DAPO-math-17k dataset \citep{yu2026dapo}, and code split of \texttt{guru-RL-92k} \citep{cheng2026revisiting}. 
For each problem, we generate four responses from both the student and advisor models and compute their empirical success rates, denoted by $r_w$ and $r_s$, respectively. 
Each problem is assigned one of three labels: advisor win if $r_s > r_w$, tie if $r_s = r_w$, and student win otherwise.
Then we fully fine-tune \texttt{ModernBERT-base} as a three-class classifier.
To mitigate label imbalance, we use inverse-frequency class weights.
At inference time, we use the predicted advisor-win probability as the routing score,
\begin{align}
    p_{\mathrm{adv}}=1-(p_{\mathrm{tie}}+p_{\mathrm{student}}).
\end{align}
A query is routed to the advisor when $p_{\mathrm{adv}} \ge \tau$.
We obtain different operating points by varying $\tau$.

\paragraph{TRIM.} For TRIM, we used TRIM-Thr method, which is a training-free step-level routing baseline that uses a fixed PRM threshold to decide whether each student-generated reasoning step should be accepted or regenerated by the advisor.
At each step, the SLM first proposes a reasoning step, which is scored by math PRM, \texttt{Qwen2.5-Math-PRM-7B} \citep{zhang2025lessons}. 
If the PRM score falls below a threshold $\rho$, the proposed step is discarded and regenerated by the LLM.
We used the double newlines as step boundaries, with at most 30 reasoning steps. 
We evaluate thresholds $\rho\in\{0.3,0.5,0.7,0.9\}$.
Additionally, we calculated the ratio of input and output token cost in \Cref{fig:TRIM}.
As shown in the figure, the advisor cost of TRIM-Thr is dominated by input tokens rather than generated output tokens. 
This is because each escalation re-sends the entire accepted prefix to the advisor, causing the input cost to grow rapidly as the routing threshold increases. 
Across benchmarks, increasing the thresholds $\rho$ from 0.3 to 0.9 raises the advisor cost by roughly $3.2$--$3.7\times$, while the input-to-output token ratio reaches up to $39\times$.
\input{figures/app-steer}

\paragraph{STEER.} Following the original method, STEER estimates step-level confidence directly from the model logits and fits a two-component Gaussian mixture model (GMM) over the confidence scores of ongoing reasoning traces. 
The posterior probability of belonging to the confident component is then thresholded to determine whether the next reasoning step is generated by the student or the advisor.
We use the original configuration for mathematical reasoning, including maximum-logit-based confidence aggregated over mathematical tokens and GMM-responsibility-based routing.
We sweep the GMM threshold over $\{0.2,0.3,0.4,0.5,0.6\}$.
We also calculated the ratio of input and output token cost for STEER \Cref{fig:STEER}.
Similar to TRIM, the input token cost accounts for a large portion of the total cost, although this tendency is relatively less pronounced.

\input{figures/app-cost-sweep-math}
\input{figures/app-cost-sweep-code}
\input{figures/app-transfer-nothink}
\input{figures/app-transfer-think}

%% file: figures/app-trim.tex
\begin{figure}[t]
    \centering
    \includegraphics[
        width=\linewidth
    ]{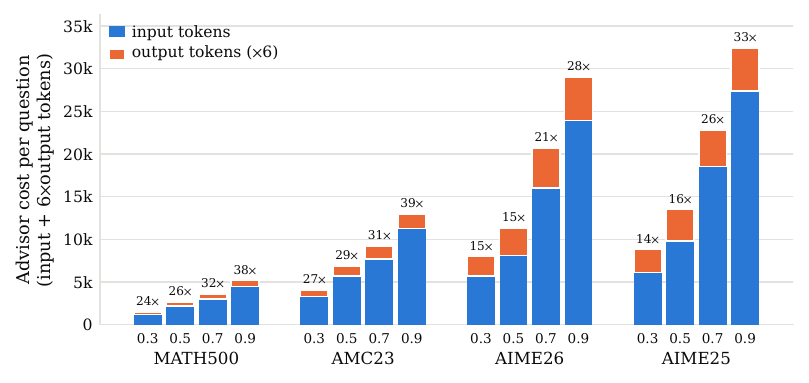}
    \caption{
    \textbf{Advisor token cost of TRIM-Thr across routing thresholds.}
    Each bar shows the average advisor cost per question, decomposed into input tokens and $6\times$-weighted output tokens.
    Numbers above bars denote the input/output token ratio, showing that advisor cost is dominated by repeated input-prefix processing, especially at higher routing thresholds.
    }
    \label{fig:TRIM}
\end{figure}

%% file: figures/app-steer.tex
\begin{figure}[t]
    \centering
    \includegraphics[
        width=\linewidth
    ]{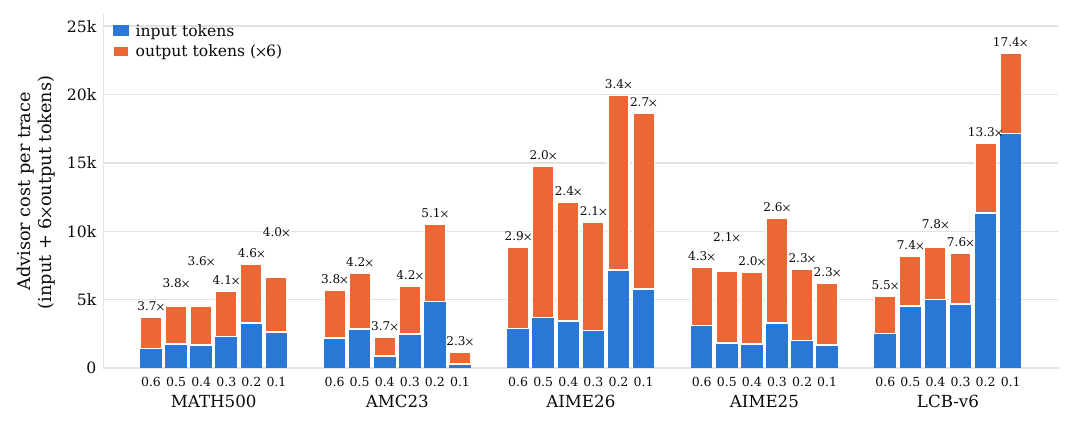}
    \caption{
    \textbf{Advisor token cost of STEER across routing thresholds.}
    Each bar shows the average advisor cost per question, decomposed into input tokens and $6\times$-weighted output tokens.
    Numbers above bars denote the input/output token ratio. The input cost accounts for a large portion of the total cost.
    }
    \label{fig:STEER}
\end{figure}

%% file: figures/app-cost-sweep-math.tex
\begin{figure}[t]
    \centering
    \includegraphics[
        width=\linewidth
    ]{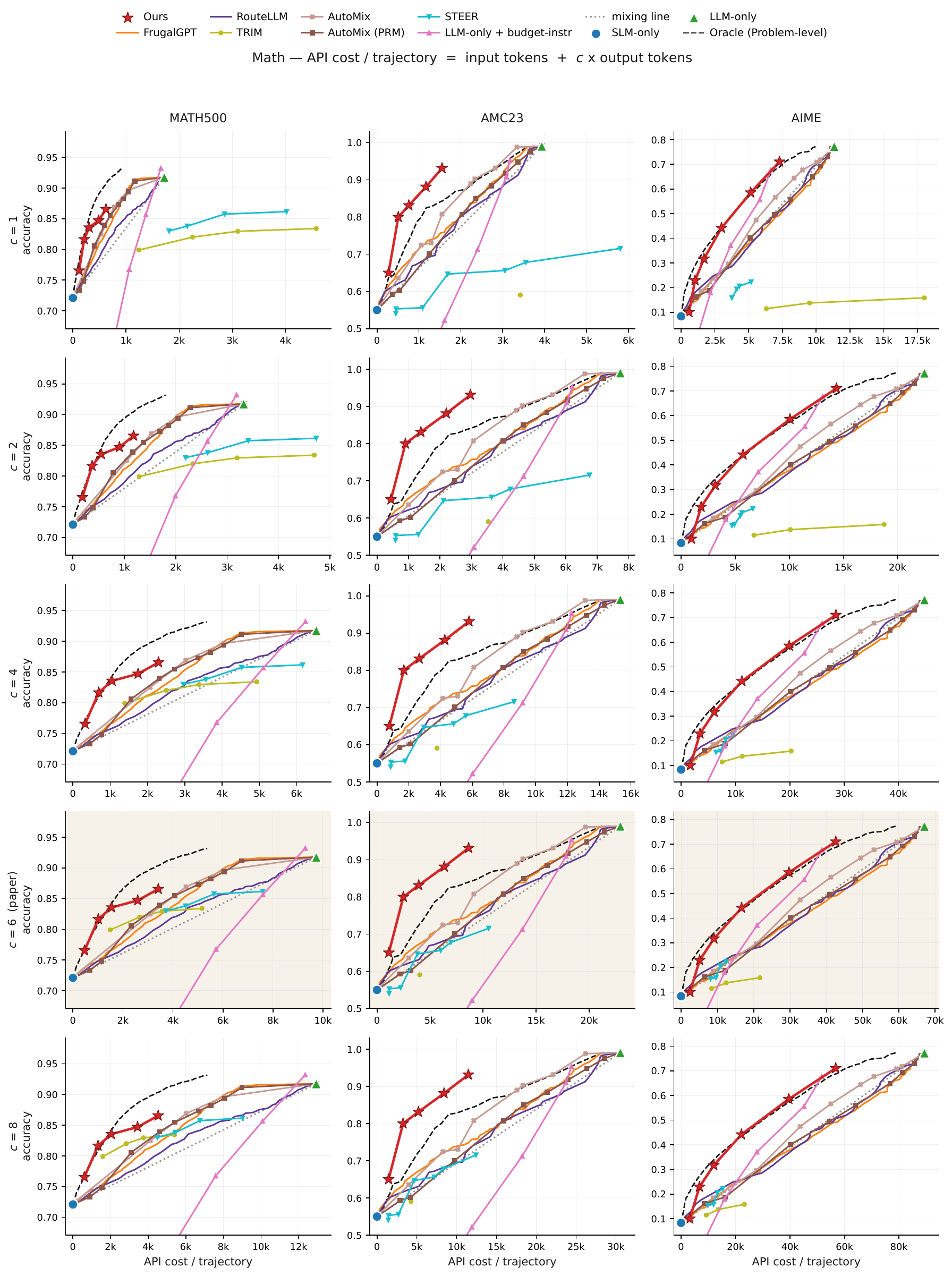}
    \caption{
    \textbf{The performance-cost pareto front for mathmatics benchmark.}
    }
    \label{fig:cost-sweep-math}
\end{figure}

%% file: figures/app-cost-sweep-code.tex
\begin{figure}[t]
    \centering
    \includegraphics[
        width=\linewidth
    ]{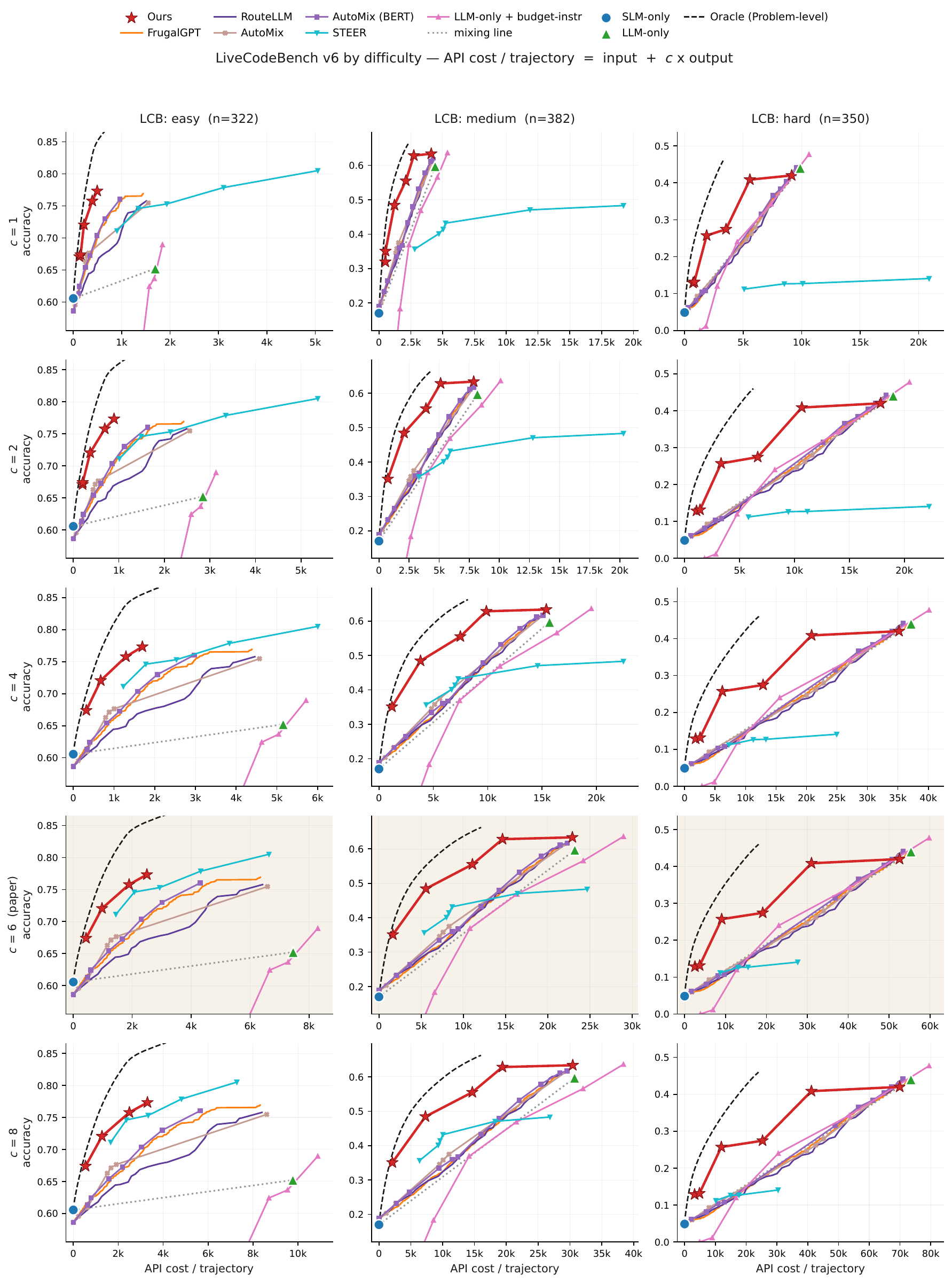}
    \caption{
    \textbf{The performance-cost pareto front for LiveCodeBench v6, across different difficulty group.}
    }
    \label{fig:cost-sweep-code}
\end{figure}

%% file: figures/app-transfer-nothink.tex
\begin{figure}[t]
    \centering
    \includegraphics[
        width=\linewidth
    ]{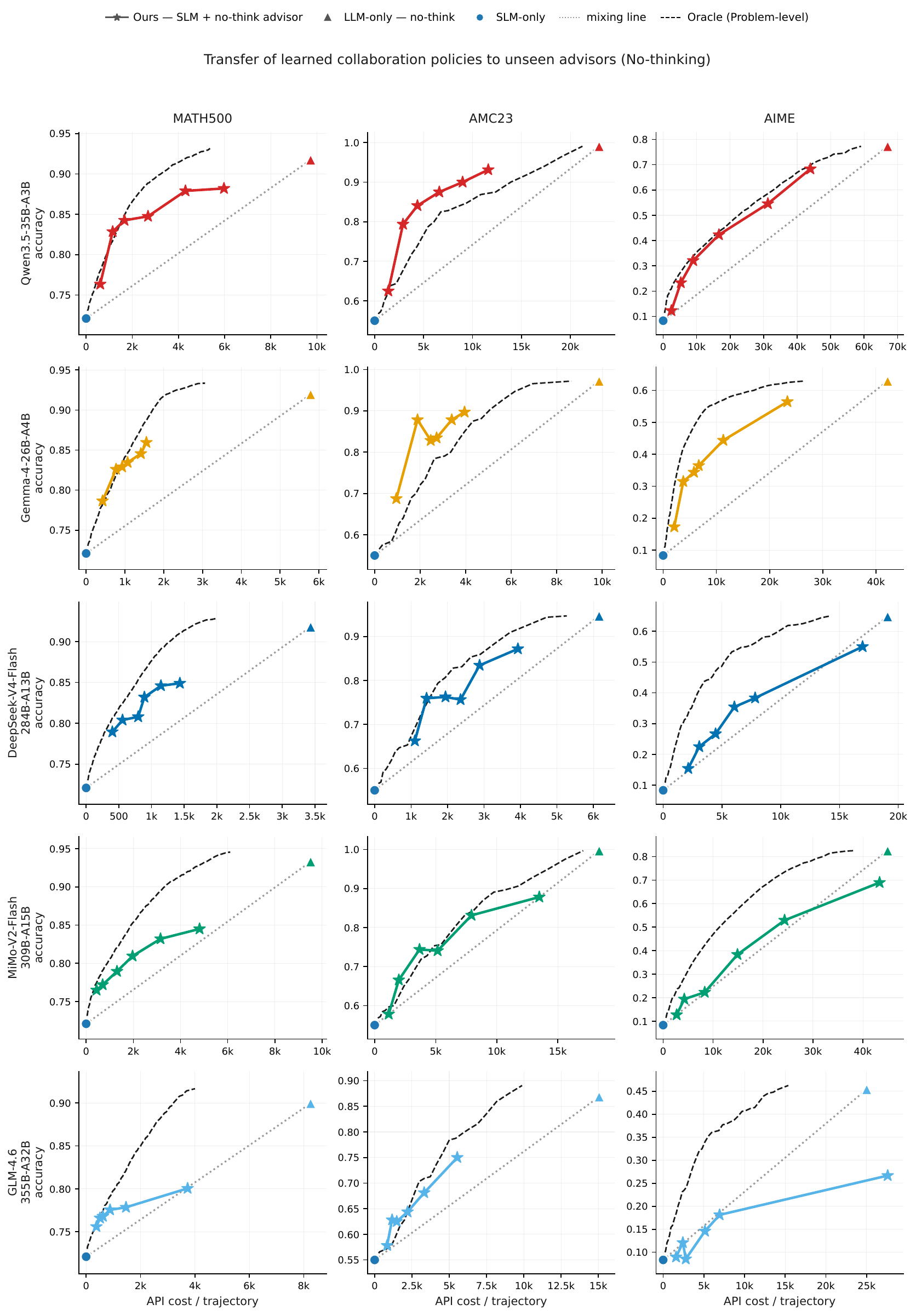}
    \caption{
    \textbf{Transfer to unseen LLM advisors.} We evaluate whether a collaboration policy trained with one LLM advisor transfers to unseen advisor models.
    }
    \label{fig:transfer-no-think}
\end{figure}

%% file: figures/app-transfer-think.tex
\begin{figure}[t]
    \centering
    \includegraphics[
        width=\linewidth
    ]{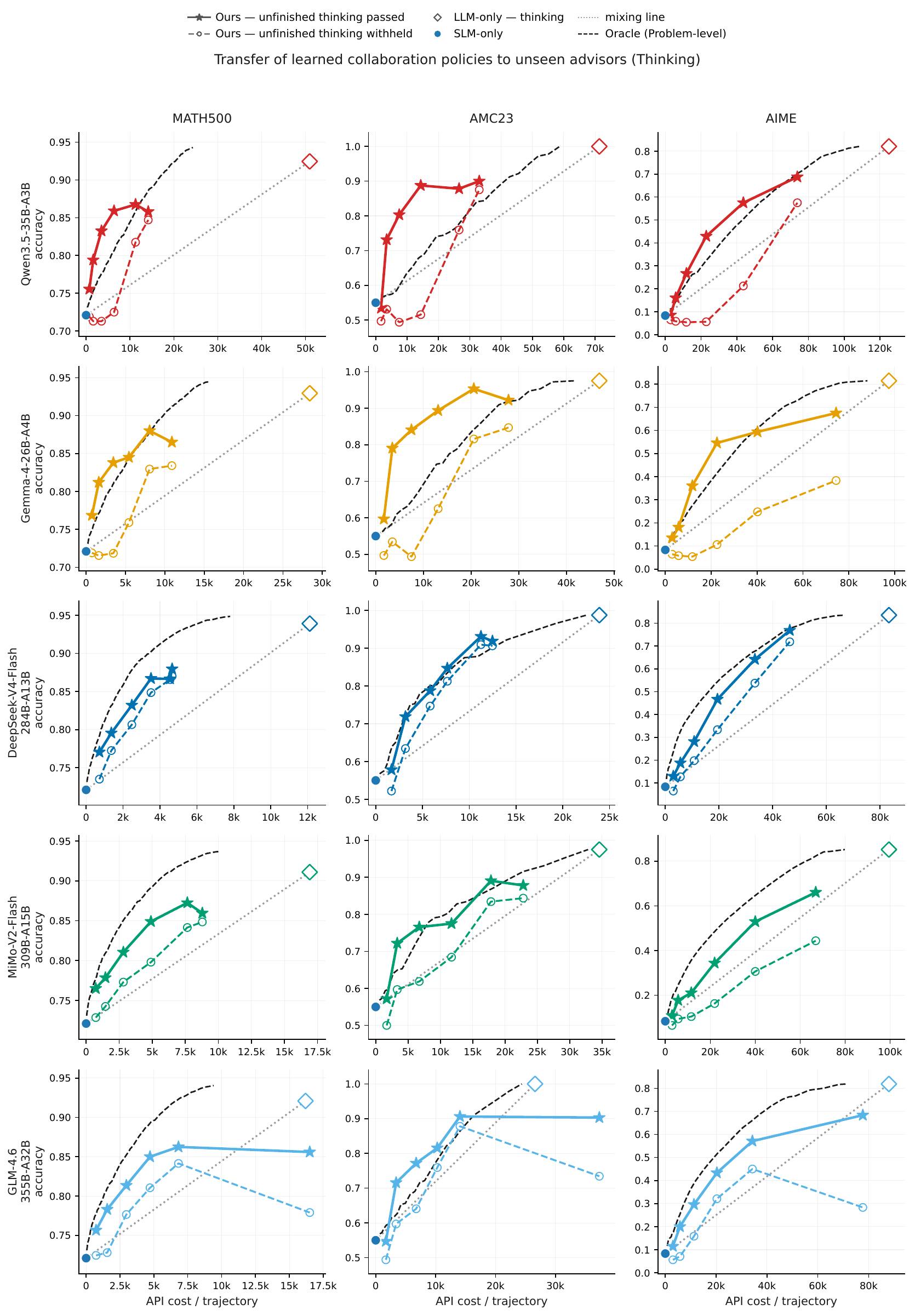}
    \caption{
    \textbf{Transfer to unseen LLM advisors, on thinking mode.} 
    We evaluate whether a collaboration policy trained with one LLM advisor transfers to unseen advisor models, on thinking mode.
    We consider two transfer settings depending on whether the advisor’s unfinished reasoning is passed to or withheld from the SLM.
    }
    \label{fig:transfer-think}
\end{figure}

%% file: section/appendix/app-prompts.tex
\clearpage
\section{Prompts}

\subsection{SLM prompts}
We use task-specific system prompts for mathematical reasoning and code generation.
The mathematical prompt instructs the model to return a final answer in boxed form, whereas the code prompt requires a complete Python program as the final response.
Both prompts allow the model to query the advisor when an intermediate fact or sub-result is needed.
In the solo setting, all advisor-related instructions are removed from the system prompt.





\begin{systempromptboxmath}

\promptlabel{System Prompt}
\begin{innerprompt}
You are a careful problem solver. Think step by step.

If you reach a point where a specific fact, definition, lemma, or sub-result would let you make progress, you may call the `ask_advisor` tool with a focused question. The advisor will respond, and you can then continue your reasoning using their answer.

When you have the final answer, wrap it in \\boxed{...}.
\end{innerprompt}
\end{systempromptboxmath}

\begin{systempromptboxcode}

\promptlabel{System Prompt}
\begin{innerprompt}
You are a careful problem solver. Think step by step.

If you reach a point where a specific fact, definition, lemma, or sub-result would let you make progress, you may call the `ask_advisor` tool with a focused question. The advisor will respond, and you can then continue your reasoning using their answer.

When you have the final solution, write the complete Python program and enclose your code within delimiters:
```python
# your code here
```
The program is your final answer --- do not wrap it in \\boxed{...} and do not fabricate example output.
\end{innerprompt}
\end{systempromptboxcode}

\subsection{Qmaker prompt}
\begin{promptboxqmaker}
    \promptlabel{System Prompt}
\begin{innerprompt}
You are a tutoring coordinator. A student is solving a problem and has paused to ask an expert advisor for help. Your job is to write (1) an analysis of where the student is stuck and what is missing, (2) a single, sharply-targeted question for the advisor based on the analysis, and (3) an instruction about reply budget.

You will see:
1. The original problem the student is solving.
2. The student's reasoning so far, ending at the moment they invoked the expert advisor.

The advisor's reply is hard-truncated by a fixed token budget for this call. A truncated reply is usually useless to the student, so a compact, well-fitted budget instruction is required.

Rules for the analysis:
- 2-5 full sentences in an <analysis> block, written FIRST: where exactly the student is stuck, what single piece of information is missing (including any intermediate result the student has already derived that should be carried into the question), and how much reply this needs.
- Do NOT solve the problem in the analysis. The analysis is for you alone --- the advisor never sees it.
- The question and the budget instruction must follow from your analysis.

Rules for the question:
- It must be self-contained. The advisor will NOT see the original problem or the student's reasoning --- only your question and your format instruction.
- Target the precise sub-problem the student is stuck on. Ask for the cheapest piece of information that lets the student continue --- exactly what is missing, nothing more.
- If the student's reasoning contains a mistake, do not build your question on top of it: state the corrected premise, or make that step itself what you ask about.
- If the student has made no progress yet (the reasoning before the call is empty or contains no actual work), do not invent a sub-problem they never reached: ask the advisor for a rough solution plan --- the key idea and the main steps to follow, not the worked solution.
- Do NOT solve the original problem yourself, and do NOT include the final answer.

Rules for the reply-budget instruction:
- One sentence, addressed directly to the advisor. It is appended verbatim after your question, so write it as a standalone imperative.
- It controls only the budget of the reply. All problem content (equations, code, facts) belongs in the question, none in the instruction.
- Choose the small but sufficient amount in which this question can still be answered CORRECTLY. Judge what this question actually needs (e.g. some number of plain prose lines) --- and ask within that much.

Output exactly three blocks and nothing else:
<analysis>...</analysis>
<question>...</question>
<instruction>...</instruction>
\end{innerprompt}

\promptlabel{User Prompt}
\begin{innerprompt}
Original problem: {problem}

Student reasoning so far: {student_prefix_text} 
\end{innerprompt}

\end{promptboxqmaker}

\subsection{Prompt for LLM}

For the LLM prompt, we use two prompts.
One is for judge the incompleteness during stage-2 training, and one is to answer the issued question from the \textit{Qmaker}.
We state both prompt as below.
Note that we use the judge of the training same as the target advisor i.e., 4-bit quantized Qwen3.5-35B-A3B with no thinking mode, different with judge that we use in analysis.

\begin{promptboxadvisor}
\promptlabel{System Prompt}
\begin{innerprompt}
You are a strict screener. Follow the instructions exactly.

\end{innerprompt}
\promptlabel{User Prompt}
\begin{innerprompt}
You are screening a question that a student wants to send to an expert advisor. The advisor will see ONLY this question --- no other context: not the original problem, not the student's work.

QUESTION:
{Question from the Qmaker}

Decide whether the question is SELF-CONTAINED: answerable exactly as written, on its own. It is INCOMPLETE if it refers to context the advisor cannot see ("the equation above", "this function", "the value we found", "the problem", "the given conditions"), omits a definition, value, or constraint needed to answer it, is a bare number or expression with nothing actually asked, or is otherwise ambiguous without outside context.

If INCOMPLETE, reply with EXACTLY this line and nothing else:
It is not valid question.

Otherwise reply with EXACTLY:
VALID
\end{innerprompt}

\end{promptboxadvisor}

\begin{promptboxadvisor}
\promptlabel{System Prompt}
\begin{innerprompt}
You are an expert advisor helping a student make progress on a problem they are stuck on. The message you receive has two parts: a self-contained question, followed by a reply-format instruction.

Answer the question exactly as asked, accurately and self-containedly. Follow the reply-format instruction strictly --- match the requested length and form of the reply. Do not ask clarifying questions --- give your best answer based on the question as written.
\end{innerprompt}
\promptlabel{User Prompt}
\begin{innerprompt}
    {Question from the Qmaker}

    {Instruction from the Qmaker}
\end{innerprompt}

\end{promptboxadvisor}

\subsection{Tool description}
\begin{tooldescriptionbox}
ASK_ADVISOR_TOOL_SCHEMA: dict = {
    "type": "function",
    "function": {
        "name": ASK_ADVISOR_TOOL_NAME,
        "description": (
            "Ask an expert advisor one focused question to make progress on the current problem. Ask for the specific thing you are missing at this point in your reasoning. The advisor sees ONLY your question --- not the original problem, not your reasoning so far, and no other context. Anything the advisor needs to know must be stated inside the question itself."
        ),
        "parameters": {
            "type": "object",
            "properties": {
                "query": {
                    "type": "string",
                    "description": "A self-contained question to ask the expert advisor.",
                },
            },
            "required": ["query"],
        },
    },
}
\end{tooldescriptionbox}

\subsection{Judge Prompt}
\label{app:judge-prompt}
We use an LLM judge to classify each query (Q2) with respect to its corresponding original problem (Q1). 
The judge applies a sequence of checks in a fixed order, distinguishing incomplete questions, transformed questions, questions containing additional information or hints, and restatements of the full task.
The exact prompt used for classification is shown below.
For all queries, we used GPT-5.6 Luna \citep{openai2026gpt56} as the LLM judge, with medium level of reasoning effort.
All other hyperparameters are set to default.

\begin{promptboxjudge}
    Q1:
{q1}

Q2:
{q2}

Q1 is an original problem. Q2 is a question someone wrote while working on Q1.

Classify Q2 by applying the following checks IN ORDER and stopping at the
first that decides:

Step 1 --- INCOMPLETE: Q2 alone cannot be answered as written: it refers to context not stated in Q2, or omits a definition, value, or constraint needed to answer it. If so, reply INCOMPLETE.

Step 2 --- SAME QUESTION: check whether Q2 asks exactly the same thing as Q1, without transforming the content. Copies, paraphrases, and formatting/LaTeX changes count as the same.
If what Q2 asks is not exactly what Q1 asks --- any transformation exists: it asks for only a piece or an intermediate result, asks for the method instead of the answer, asks for a different final form or target, or asks something else entirely --- reply TRANSFORM.

Step 3 --- ADDED INFORMATION: check whether Q2 states anything that is not stated in Q1 --- derived values, intermediate results, hints, examples absent from Q1, narrowed or corrected assumptions. If Q2 adds information, reply HINT. If Q2 adds nothing, reply FULL_TASK.

Reply with EXACTLY one word: INCOMPLETE, TRANSFORM, HINT, or FULL_TASK
\end{promptboxjudge}

\subsection{Sensitive text filtering prompt}
\label{app:sensitive-prompt}

\begin{promptboxqmaker}
    \promptlabel{System Prompt}
\begin{innerprompt}
You are a tutoring coordinator. A student is solving a problem and has paused to ask an expert advisor for help. Your job is to write (1) an analysis of where the student is stuck and what is missing, (2) a single, sharply-targeted question for the advisor based on the analysis, and (3) an instruction about reply budget.

... (Same prompt with normal Qmaker) ...

Privacy rules (sensitive values):
- The problem or the student's work may contain sensitive values that must not leave the local environment --- credentials and secrets such as API keys or passwords, and similar private values, even in forms not listed here.
- Your <analysis> block is local: you may reference sensitive values there freely.
- Your <question> and <instruction> blocks are sent externally: they must never contain a sensitive value. If the task refers to one, replace it with a consistent neutral placeholder (e.g. <API_KEY>, <PASSWORD>) so the question still reads coherently and the advisor's answer remains usable after the placeholder is substituted back locally.

Output exactly three blocks and nothing else:
<analysis>...</analysis>
<question>...</question>
<instruction>...</instruction>
\end{innerprompt}

\promptlabel{User Prompt}
\begin{innerprompt}
Original problem: {problem}

Student reasoning so far: {student_prefix_text} 
\end{innerprompt}

\end{promptboxqmaker}

\begin{systempromptfilter}
\promptlabel{System Prompt}
\begin{innerprompt}
You are a local privacy filter. You are given the full text of a programming task that is about to be sent to an external model.

Rewrite the text so that no sensitive value leaves the local environment. Sensitive values include: passwords, API keys, access tokens, internal service URLs, and other designated secret values.

Rules:
- Replace every sensitive VALUE with the literal token PLACEHOLDER. Keep the surrounding sentence, variable name and formatting intact so the task still reads naturally.
- Preserve everything else exactly as written: the problem statement, constraints, examples, input/output format, and any requirement that mentions the value (the external model must still know that a value has to be printed or returned there).
- Do NOT solve the task, do not add commentary, do not summarise, do not shorten.
- Output only the rewritten task text.
\end{innerprompt}
\end{systempromptfilter}

%% file: section/appendix/app-gen-examples.tex
\section{Generation examples}
\label{app:generation-examples}
Here, we present illustrative generation examples from the trained collaboration pipeline. 
The examples illustrate how the SLM selectively queries the advisor at different levels of granularity, ranging from targeted subproblems and reasoning hints to full-task requests when necessary.
They also show how the SLM incorporates the returned information into its subsequent reasoning, often completing substantial parts of the solution itself. 
Together, these examples show the diverse collaboration patterns that emerge under different communication budgets.

\begin{generationexample}{
    MATH500, Problem 475
    \hfill
    \textnormal{Advisor budget: 512 tokens}
}

\begin{problembox}
When rolling a certain unfair six-sided die with faces numbered
$1,2,3,4,5,$ and $6$, the probability of obtaining face $F$ is
greater than $1/6$, the probability of obtaining the face opposite
face $F$ is less than $1/6$, the probability of obtaining each of the
other faces is $1/6$, and the sum of the numbers on each pair of
opposite faces is $7$. When two such dice are rolled, the probability
of obtaining a sum of $7$ is $\frac{47}{288}$. Given that the
probability of obtaining face $F$ is $\frac{m}{n}$, where $m$ and $n$
are relatively prime positive integers, find $m+n$.
\end{problembox}

Let the probability of obtaining face $F$ be $p$, and let the
probability of obtaining its opposite face be $q$. Since each of the
other four faces has probability $\frac16$,
\[
p+q+4\cdot\frac16=1,
\]
so
\[
p+q=\frac13.
\]
The probability of obtaining a sum of $7$ when rolling two such dice is
$\frac{47}{288}$. We need to express this probability in terms of $p$
and $q$ and solve for the probability of face $F$.

\begin{qmakerbox}
\qmakerrequest{
Given a six-sided die with faces numbered 1, 2, 3, 4, 5, and 6, where
the probability of obtaining face $F$ is $p$, the probability of
obtaining the face opposite face $F$ is $q$, and the probability of
obtaining each of the other faces is $\frac{1}{6}$. The sum of the
numbers on each pair of opposite faces is 7. If the probability of
obtaining a sum of 7 when two such dice are rolled is
$\frac{47}{288}$, find the value of $p$ and $q$ such that
$p+q=\frac{1}{3}$.
}

\qmakerbudget{
Answer in 5--7 lines.
}
\end{qmakerbox}

\begin{advisorbox}
Let the face $F$ and its opposite $7-F$ have probabilities $p$ and
$q$, while the other four faces each have probability $\frac16$.
The pairs summing to $7$ involving the two biased faces are
$(F,7-F)$ and $(7-F,F)$, contributing $2pq$.

The remaining four neutral faces form two opposite pairs. Their total
contribution is
\[
4\left(\frac16\right)^2=\frac19.
\]
Hence the probability of obtaining a sum of $7$ is
\[
2pq+\frac19.
\]

\end{advisorbox}

Using the advisor's relation,
\[
2pq+\frac19=\frac{47}{288}.
\]
Since
\[
q=\frac13-p,
\]
we obtain
\[
2p\left(\frac13-p\right)+\frac19=\frac{47}{288}.
\]
Multiplying through by $288$ and simplifying gives
\[
576p^2-192p+15=0.
\]
Thus,
\[
p
=\frac{192\pm\sqrt{192^2-4(576)(15)}}{1152}
=\frac{192\pm48}{1152},
\]
so
\[
p=\frac{5}{24}
\qquad\text{or}\qquad
p=\frac18.
\]
Because the problem states that the probability of face $F$ is greater
than $\frac16$, we must have
\[
p=\frac{5}{24},
\qquad
q=\frac18.
\]
Therefore, if $p=\frac{m}{n}$ in lowest terms,
\[
m+n=5+24=\boxed{29}.
\]

\interpretation[Full-task]{
The advisor identifies the key probability decomposition
$2pq+\frac19$ but does not complete the algebra. The SLM independently
substitutes $p+q=\frac13$, derives and solves the resulting quadratic,
uses the original inequality $p>\frac16$ to disambiguate the two roots,
and performs the final answer transformation. This example illustrates
that the advisor response can serve as intermediate scaffolding while
the SLM retains substantial responsibility for completing the solution.
}

\end{generationexample}

\begin{generationexample}{
    AMC23, 24 
    \hfill
    \textnormal{Advisor budget: 512 tokens}
}

\begin{problembox}
Calculate the maximum area of an isosceles trapezoid that has legs of length $1$ and one base twice as long as the other. The final answer can be written in the form $\frac{m}{n}$, where $m$ and $n$ are relatively prime positive integers. What is $m^2+n^2$?

\end{problembox}
To find the maximum area of an isosceles trapezoid with legs of length 1 and one base twice as long as the other, we start by defining the lengths of the bases. Let the shorter base be \(a\) and the longer base be \(2a\). The height of the trapezoid is denoted by \(h\).

We can split the trapezoid into a rectangle and two right triangles. The right triangles are congruent, and each has a leg of length \(h\) and the other leg of length \(\frac{2a - a}{2} = \frac{a}{2}\). The hypotenuse of each right triangle is the leg of the trapezoid, which is 1. Therefore, we can use the Pythagorean theorem to find the height \(h\):

\[
1^2 = h^2 + \left(\frac{a}{2}\right)^2
\]

This simplifies to:

\[
1 = h^2 + \frac{a^2}{4}
\]

Solving for \(h^2\), we get:

\[
h^2 = 1 - \frac{a^2}{4}
\]

The area \(A\) of the trapezoid is given by the formula:

\[
A = \frac{1}{2} \times (a + 2a) \times h = \frac{3a}{2} \times h
\]

Substituting \(h^2 = 1 - \frac{a^2}{4}\) into the area formula, we need to maximize \(A\). To do this, we express \(h\) in terms of \(a\):

\[
h = \sqrt{1 - \frac{a^2}{4}}
\]

Thus, the area \(A\) becomes:

\[
A = \frac{3a}{2} \sqrt{1 - \frac{a^2}{4}}
\]

To find the maximum value of \(A\), we can use calculus or recognize that the expression inside the square root and the linear term in \(a\) suggest a maximum value when the derivative of \(A\) with respect to \(a\) is zero. However, a simpler approach is to use the advisor to find the critical value of \(a\) that maximizes the area.

We call the advisor to find the specific value of \(a\) that maximizes the area.

\begin{qmakerbox}
\qmakerrequest{
An isosceles trapezoid has legs of length 1 and one base twice as long as the other. Find the value of \( a \) that maximizes the area of the trapezoid, given the expression for the area \( A = \frac{3a}{2} \sqrt{1 - \frac{a^2}{4}} \).

}

\qmakerbudget{
Reply with 5-7 lines of prose.

}
\end{qmakerbox}

\begin{advisorbox}

To maximize the area, we first square the given expression to simplify the optimization process, yielding $A^2 = \frac{9a^2}{4} \left(1 - \frac{a^2}{4}\right)$. Letting $x = a^2$, the function becomes a downward-opening parabola $f(x) = \frac{9}{4}(x - \frac{x^2}{4})$. The vertex of this parabola occurs at $x = \frac{-1}{2(-1/4)} = 2$, which corresponds to $a^2 = 2$. Solving for $a$, we find that the positive value maximizing the area is $\sqrt{2}$. This result is valid because $a = \sqrt{2}$ lies within the domain $0 < a < 2$ required for the square root to be real. Thus, the specific value of $a$ that yields the maximum area is $\sqrt{2}$.

\end{advisorbox}
Given that the value of \(a\) that maximizes the area is \(\sqrt{2}\), we substitute \(a = \sqrt{2}\) back into the expression for the height \(h\):

\[
h = \sqrt{1 - \frac{(\sqrt{2})^2}{4}} = \sqrt{1 - \frac{2}{4}} = \sqrt{1 - \frac{1}{2}} = \sqrt{\frac{1}{2}} = \frac{1}{\sqrt{2}} = \frac{\sqrt{2}}{2}
\]

Now, we calculate the area \(A\) using the formula for the area of the trapezoid:

\[
A = \frac{3a}{2} \times h = \frac{3\sqrt{2}}{2} \times \frac{\sqrt{2}}{2} = \frac{3\sqrt{2} \cdot \sqrt{2}}{4} = \frac{3 \cdot 2}{4} = \frac{6}{4} = \frac{3}{2}
\]

The area of the trapezoid is \(\frac{3}{2}\). In the form \(\frac{m}{n}\), we have \(m = 3\) and \(n = 2\). Therefore, \(m^2 + n^2 = 3^2 + 2^2 = 9 + 4 = 13\).

The final answer is:

\[
\boxed{13}
\]

\interpretation[Transform]{
The SLM first derives the geometric model on its own, expressing the height and the area.
It then delegates only the optimization subproblem, asking the advisor to determine the value of $a$ that maximizes this expression. The advisor solves this exactly, and the SLM then resumes the solution, substitutes this value back into the height and area formulas, obtains the final correct solution. Thus, the advisor supplies only the critical maximizing value, while the SLM is responsible for both constructing the objective function and completing the original task.
}

\end{generationexample}

\begin{generationexample}{
    AMC23, 13 
    \hfill
    \textnormal{Advisor budget: 512 tokens}
}

\begin{problembox}
How many ordered pairs of positive real numbers $(a,b)$ satisfy the equation
\[(1+2a)(2+2b)(2a+b) = 32ab?\]
\end{problembox}
To solve the equation \((1+2a)(2+2b)(2a+b) = 32ab\) for ordered pairs of positive real numbers \((a, b)\), we start by expanding and simplifying the left-hand side.

First, let's rewrite the equation:
\[
(1+2a)(2+2b)(2a+b) = 32ab.
\]

We will use the advisor to get a specific insight into how to proceed with this equation. Let's call the advisor with a focused question about the structure of the equation.

\begin{qmakerbox}
\qmakerrequest{
Given the equation \((1+2a)(2+2b)(2a+b) = 32ab\) for positive real numbers \(a\) and \(b\), what specific fact, definition, lemma, or sub-result would help make progress in solving for \(a\) and \(b\)?

}

\qmakerbudget{
Answer in 5-7 lines.

}
\end{qmakerbox}

\begin{advisorbox}
The most helpful approach is to apply the **Arithmetic Mean-Geometric Mean (AM-GM) inequality** to each of the three factors on the left-hand side.
For positive reals, we know that $(1+2a) \ge 2\sqrt{2a}$, $(2+2b) \ge 4\sqrt{b}$, and $(2a+b) \ge 2\sqrt{2ab}$.
Multiplying these three inequalities yields a lower bound for the product that simplifies directly to $32ab$.
Since the given equation states the product equals exactly $32ab$, the equality condition for AM-GM must hold simultaneously for all three factors.
This specific fact forces the equalities $1=2a$, $2=2b$, and $2a=b$ to be true at the same time.
Solving these resulting simple linear equations immediately determines the unique values for $a$ and $b$.
Thus, the equality case of the AM-GM inequality is the critical lemma that reduces the complex equation to a solvable system.

\end{advisorbox}
Given the advisor's guidance, we need to apply the Arithmetic Mean-Geometric Mean (AM-GM) inequality to each of the three factors on the left-hand side of the equation \((1+2a)(2+2b)(2a+b) = 32ab\).

The AM-GM inequality states that for non-negative real numbers \(x_1, x_2, \ldots, x_n\),
\[
\frac{x_1 + x_2 + \cdots + x_n}{n} \ge \sqrt[n]{x_1 x_2 \cdots x_n},
\]
with equality if and only if \(x_1 = x_2 = \cdots = x_n\).

Applying AM-GM to each factor:
1. For \(1 + 2a\):
\[
1 + 2a \ge 2\sqrt{2a}.
\]
Equality holds when \(1 = 2a\), so \(a = \frac{1}{2}\).

2. For \(2 + 2b\):
\[
2 + 2b \ge 4\sqrt{b}.
\]
Equality holds when \(2 = 2b\), so \(b = 1\).

3. For \(2a + b\):
\[
2a + b \ge 2\sqrt{2ab}.
\]
Equality holds when \(2a = b\), which we already know from the previous two conditions.

Substituting \(a = \frac{1}{2}\) and \(b = 1\) into the equation, we check:
\[
(1 + 2 \cdot \frac{1}{2})(2 + 2 \cdot 1)(2 \cdot \frac{1}{2} + 1) = (1 + 1)(2 + 2)(1 + 1) = 2 \cdot 4 \cdot 2 = 16.
\]
However, we need to verify if this satisfies the original equation:
\[
(1 + 2a)(2 + 2b)(2a + b) = 32ab.
\]
Substituting \(a = \frac{1}{2}\) and \(b = 1\):
\[
(1 + 2 \cdot \frac{1}{2})(2 + 2 \cdot 1)(2 \cdot \frac{1}{2} + 1) = 32 \cdot \frac{1}{2} \cdot 1.
\]
This simplifies to:
\[
2 \cdot 4 \cdot 2 = 32 \cdot \frac{1}{2} = 16.
\]
Both sides are equal, confirming that \((a, b) = \left(\frac{1}{2}, 1\right)\) is a solution.

Since the equality condition of AM-GM uniquely determines \(a\) and \(b\), there is only one solution. Therefore, the number of ordered pairs \((a, b)\) that satisfy the equation is:
\[
\boxed{1}.
\]

\interpretation[Transform]{
    The SLM asks for a useful fact, lemma, or intermediate result that would make progress possible.
    The advisor identifies the AM-GM inequality as the key tool, gives the three relevant lower bounds, and states the simultaneous equality conditions. Using that guidance, the SLM carries out the remainder of the argument itself, leading to the correct answer.
}

\end{generationexample}

\begin{generationexample}{
    MATH500, 434 
    \hfill 
    \textnormal{Advisor budget: 512 tokens}
}

\begin{problembox}
In a certain isosceles right triangle, the altitude to the hypotenuse has length $4\sqrt{2}$.  What is the area of the triangle?

\end{problembox}
To find the area of the isosceles right triangle given that the altitude to the hypotenuse is \(4\sqrt{2}\), we can follow these steps:

1. **Identify the properties of the isosceles right triangle:**
   - In an isosceles right triangle, the two legs are equal, and the hypotenuse is \( \sqrt{2} \) times the length of each leg.
   - Let the legs of the triangle be \(a\). Then the hypotenuse is \(a\sqrt{2}\).

2. **Use the formula for the area of a triangle:**
   - The area \(A\) of a triangle can be expressed as \( \frac{1}{2} \times \text{base} \times \text{height} \).
   - Here, the base is the hypotenuse \(a\sqrt{2}\) and the height is the altitude to the hypotenuse, which is given as \(4\sqrt{2}\).

3. **Set up the area equation:**
   \[
   A = \frac{1}{2} \times a\sqrt{2} \times 4\sqrt{2}
   \]
   Simplify the expression:
   \[
   A = \frac{1}{2} \times a\sqrt{2} \times 4\sqrt{2} = \frac{1}{2} \times a \times 8 = 4a
   \]

4. **Relate the altitude to the hypotenuse:**
   - The altitude to the hypotenuse in an isosceles right triangle can also be found using the relationship involving the legs and the hypotenuse. However, we need to find the exact value of \(a\) first.

5. **Use the advisor to find the specific relationship:**
   Since we have simplified the problem to finding the value of \(a\), we can call the advisor for further assistance.

\begin{qmakerbox}
\qmakerrequest{
Given an isosceles right triangle with an altitude to the hypotenuse of length \(4\sqrt{2}\), how can we find the length of the legs of the triangle?
}

\qmakerbudget{
Reply with 5-7 lines of prose.
}
\end{qmakerbox}

\begin{advisorbox}

In an isosceles right triangle, the altitude drawn to the hypotenuse bisects it and creates two smaller triangles that are also isosceles right triangles. Consequently, the length of this altitude is exactly equal to half the length of the hypotenuse, meaning the hypotenuse itself measures $8\sqrt{2}$. Since the triangle is isosceles, the two legs are of equal length, let's call this length $x$. Applying the Pythagorean theorem, we get $x^2 + x^2 = (8\sqrt{2})^2$, which simplifies to $2x^2 = 128$. Solving for $x$ gives $x^2 = 64$, so the length of each leg is 8.

\end{advisorbox}
Now that we know the length of each leg of the isosceles right triangle is 8, we can calculate the area of the triangle.

The area \(A\) of the triangle is given by:
\[
A = \frac{1}{2} \times \text{base} \times \text{height}
\]
Here, the base and height are both the legs of the triangle, each of length 8. Therefore,
\[
A = \frac{1}{2} \times 8 \times 8 = \frac{1}{2} \times 64 = 32
\]

Thus, the area of the triangle is \(\boxed{32}\).

\interpretation[Transform]{
The SLM first reduces the original area problem to finding the common leg length $a$. It identifies the hypotenuse and altitude as the natural base-height pair for the area formula and then asks the advisor only how to determine the legs of the isosceles right triangle from the given altitude. The advisor supplies this intermediate geometric calculation.
The SLM takes this intermediate result and finishes the original task independently
}

\end{generationexample}

\begin{generationexample}{
    AMC23, Problem 18
    \hfill
    \textnormal{Advisor budget: 512 tokens}
}
\begin{problembox}
Janet rolls a standard $6$-sided die $4$ times and keeps a running
total of the numbers she rolls. What is the probability that at some
point, her running total will equal $3$? The final answer can be
written in the form $\frac{m}{n}$, where $m$ and $n$ are relatively
prime positive integers. What is $m+n$?
\end{problembox}

To determine the probability that Janet's running total equals $3$
at some point, we need to count the roll sequences that reach a running
total of exactly $3$. We can ask the advisor for this counting step.

\begin{qmakerbox}
\qmakerrequest{
Janet rolls a standard 6-sided die 4 times and keeps a running total
of the numbers she rolls. How many sequences of 4 rolls result in a
running total of 3 at some point?
}

\qmakerbudget{
Answer in 5--7 lines.
}
\end{qmakerbox}

\begin{advisorbox}
The running total can reach $3$ on the first, second, or third roll.
If it occurs on the first roll, the first roll must be $3$, giving
$6^3=216$ continuations. If it occurs on the second roll, the first
two rolls are $(1,2)$ or $(2,1)$, giving $2\cdot6^2=72$ sequences.
If it occurs on the third roll, the first three rolls must be
$(1,1,1)$, giving $6$ possibilities for the fourth roll. Thus there
are $294$ valid sequences.
\end{advisorbox}

There are $294$ valid sequences out of $6^4=1296$ total sequences, so
\[
\frac{294}{1296}=\frac{49}{216}.
\]
Thus $m=49$ and $n=216$, giving
\[
m+n=\boxed{265}.
\]

\interpretation[Transform]{
The QMaker isolates the combinatorial counting step rather than
re-delivering the full task. The advisor supplies the intermediate
count, while the SLM retains responsibility for converting it into
a probability and producing the final answer.
}

\end{generationexample}

\begin{generationexample}{
    AIME 2025, Problem 6
    \hfill
    \textnormal{Advisor budget: 1024 tokens}
}

\begin{problembox}
An isosceles trapezoid has an inscribed circle tangent to each of its
four sides. The radius of the circle is $3$, and the area of the
trapezoid is $72$. Let the parallel sides of the trapezoid have lengths
$r$ and $s$, with $r \ne s$. Find $r^2+s^2$.
\end{problembox}

For a tangential quadrilateral, the sums of opposite side lengths are
equal. If the two equal legs of the isosceles trapezoid have length
$t$, then
\[
r+s=2t.
\]
Also, the area of a tangential quadrilateral equals the inradius times
the semiperimeter. Since the inradius is $3$ and the area is $72$, the
semiperimeter is $24$, which gives $t=12$ and therefore
\[
r+s=24.
\]
We now need another relation between $r$ and $s$.

\begin{qmakerbox}
\qmakerrequest{
An isosceles trapezoid has an inscribed circle tangent to each of its
four sides. The radius of the circle is 3, and the area of the trapezoid
is 72. Let the parallel sides of the trapezoid have lengths $r$ and $s$
with $r \neq s$. Given that $r+s=24$, find $r^2+s^2$.
}

\qmakerbudget{
Answer within 150 tokens.
}
\end{qmakerbox}

\begin{advisorbox}
Since $r+s=24$, each leg has length $12$. The height is the diameter
of the incircle, so $h=6$. Dropping perpendiculars from the shorter
base gives a right triangle with horizontal leg
$\frac{|r-s|}{2}$. Hence
\[
6^2+\left(\frac{r-s}{2}\right)^2=12^2,
\]
so $(r-s)^2=432$. Therefore
\[
r^2+s^2
=
\frac{(r+s)^2+(r-s)^2}{2}
=
\frac{576+432}{2}
=
504.
\]
\end{advisorbox}

Thus,
\[
r^2+s^2=\boxed{504}.
\]

\interpretation[Hint]{
The QMaker incorporates the SLM-derived intermediate result
$r+s=24$ into the advisor request. Rather than restarting the problem,
the advisor continues from the SLM's current progress and supplies the
remaining geometric relation.
}

\end{generationexample}


\begin{generationexample}{
    LiveCodeBench v6, Problem 398
    \hfill
    \textnormal{Advisor budget: 512 tokens}
}
\begin{problembox}
You are given a positive integer $N$.
A positive integer $K$ is called a \emph{palindromic cube number} if
both of the following conditions hold:
\begin{itemize}
\item There exists a positive integer $x$ such that $x^3=K$.
\item The decimal representation of $K$, written without leading
zeros, is a palindrome.
\end{itemize}
Find the largest palindromic cube number $K$ satisfying $K\le N$.

\par\medskip
\textit{[...Constraints, Input, Output, Examples, ...]}
\par\medskip

\end{problembox}

\begin{verbatim}
def is_palindrome(num):
    return str(num) == str(num)[::-1]

def find_max_palindromic_cube(N):
    for x in range(int(N**(1/3)), 0, -1):
        cube = x ** 3
        if cube > N:
            continue
        if is_palindrome(cube):
            return cube
\end{verbatim}

\begin{qmakerbox}
\qmakerrequest{
You are given a positive integer N. A positive integer K is defined to
be a palindromic cube number if there is a positive integer x such that
$x^3 = K$, and the decimal representation of K is a palindrome. For
example, 343 is a palindromic cube number because $7^3 = 343$ and 343
is a palindrome.

Given a positive integer N, how can you find the maximum value of a
palindromic cube number not greater than N?

You should implement a function that takes a positive integer N and
returns the maximum palindromic cube number not greater than N.
}

\qmakerbudget{
Reply within 15 lines.
}
\end{qmakerbox}

\begin{advisorbox}
Iterate through positive integers $x$ starting from $1$. For each $x$,
compute $K=x^3$. If $K>N$, stop. Otherwise, check whether the decimal
representation of $K$ is equal to its reverse, and keep the largest
valid value. This examines only $O(N^{1/3})$ candidate cubes.

\begin{verbatim}
def max_palindromic_cube(N):
    max_k = 0
    x = 1
    while True:
        k = x * x * x
        if k > N:
            break
        if str(k) == str(k)[::-1]:
            max_k = k
        x += 1
    return max_k
\end{verbatim}
\end{advisorbox}

\begin{verbatim}
def is_palindrome(num):
    return str(num) == str(num)[::-1]

def find_max_palindromic_cube(N):
    max_k = 0
    x = 1
    while True:
        k = x ** 3
        if k > N:
            break
        if is_palindrome(k):
            max_k = k
        x += 1
    return max_k

N = int(input())
print(find_max_palindromic_cube(N))
\end{verbatim}

\interpretation[Transform]{
The QMaker extracts the algorithmic search step into a self-contained
request. The advisor supplies a robust enumeration strategy, while the
SLM remains responsible for integrating that strategy into the final
program interface, attaching input-output operations, slightly modifies the code snippet.
}

\end{generationexample}


\begin{generationexample}{
    AIME 2025, Problem 27
    \hfill
    \textnormal{Advisor budget: 2048 tokens}
}

\begin{problembox}
Let $A_1A_2A_3\ldots A_{11}$ be an $11$-sided non-convex simple
polygon satisfying the following conditions:
\begin{itemize}
    \item For each integer $2\le i\le 10$, the area of
    $\triangle A_iA_1A_{i+1}$ is $1$.
    \item For each integer $2\le i\le 10$,
    \[
    \cos(\angle A_iA_1A_{i+1})=\frac{12}{13}.
    \]
    \item The perimeter of $A_1A_2A_3\ldots A_{11}$ is $20$.
\end{itemize}

Suppose
\[
A_1A_2+A_1A_{11}
=
\frac{m\sqrt{n}-p}{q},
\]
where $m,n,p,q$ are positive integers, $n$ is squarefree, and no prime
divides all three of $m,p,$ and $q$. Find
\[
m+n+p+q.
\]
\end{problembox}

To solve the problem, we need to determine the lengths of the sides
$A_1A_2$ and $A_1A_{11}$ of the 11-sided polygon given the conditions.
We will use the advisor to get the necessary geometric insight.

First, let's summarize the given information:
\begin{enumerate}
    \item For every integer $2 \le i \le 10$, the area of
    $\triangle A_iA_1A_{i+1}$ is $1$.
    \item For every integer $2 \le i \le 10$,
    $\cos(\angle A_iA_1A_{i+1})=\frac{12}{13}$.
    \item The perimeter of the 11-gon
    $A_1A_2A_3\dots A_{11}$ is $20$.
\end{enumerate}

We need to find the lengths $A_1A_2$ and $A_1A_{11}$. To do this,
we need to understand the geometric configuration and use the given
trigonometric and area information.

We will call the advisor to get a specific geometric insight that will
help us proceed.

\begin{qmakerbox}
\qmakerrequest{
Given an 11-sided non-convex simple polygon
$A_1 A_2 A_3 \ldots A_{11}$ with the following properties:
\begin{itemize}
    \item For every integer $2 \le i \le 10$, the area of
    $\triangle A_i A_1 A_{i+1}$ is $1$.
    \item For every integer $2 \le i \le 10$,
    $\cos(\angle A_i A_1 A_{i+1})=\frac{12}{13}$.
    \item The perimeter of the 11-gon
    $A_1 A_2 A_3 \dots A_{11}$ is $20$.
\end{itemize}

Find $A_1A_2+A_1A_{11}$ in the form
$\frac{m\sqrt{n}-p}{q}$, where $m,n,p$, and $q$ are positive integers,
$n$ is not divisible by the square of any prime, and no prime divides
all of $m,p$, and $q$. Determine the value of $m+n+p+q$.
}

\qmakerbudget{
Please provide a concise but correct solution within 10--15 lines.
}
\end{qmakerbox}

\begin{advisorbox}
Let $x_i=A_1A_i$ for $i=2,\dots,11$. The problem states that for
$2\le i\le10$, the area of $\triangle A_iA_1A_{i+1}$ is 1 and
$\cos\theta_i=\frac{12}{13}$, where
$\theta_i=\angle A_iA_1A_{i+1}$.

\par\medskip
\noindent\textit{[... Omitted for brevity ...]}
\par\medskip

Wait, is it possible $A_iA_{i+1}$ are not all equal?
$d_2=a,d_3=b\implies
s_1=\sqrt{a^2+b^2-k}$.
$d_3=b,d_4=a\implies
s_2=\sqrt{b^2+a^2-k}$.
Yes, all internal sides $A_iA_{i+1}$ ($i=2,\dots,10$) are equal.

So
\[
P=a+b+9\sqrt{a^2+b^2-\frac{48}{5}}=20.
\]
Let $S=a+b$ and $P_{\mathrm{prod}}=ab=26/5$.
\[
a^2+b^2=S^2-2P_{\mathrm{prod}}=S^2-
\]
\noindent\textit{[The advisor answer cut in the middle of reasoning]}
\end{advisorbox}

To proceed, we need to determine the values of $a$ and $b$ such that
the perimeter condition is satisfied. From the advisor's response, we
know that the perimeter equation is
\[
a+b+9\sqrt{a^2+b^2-\frac{48}{5}}=20.
\]

\par\medskip
\noindent\textit{[... Omitted for brevity ...]}
\par\medskip

Solving this quadratic equation will give us the values of $a$ and
$b$. However, we only need $a+b$, which we have already determined.

The final answer is
\[
a+b=\frac{-1+9\sqrt5}{4}.
\]

To find $m+n+p+q$, we need to express $a+b$ in the form
$\frac{m\sqrt n-p}{q}$. Here,
$m=9$, $n=5$, $p=1$, and $q=4$.

Thus,
\[
m+n+p+q=9+5+1+4=19.
\]

The answer is
\[
\boxed{19}.
\]

\interpretation[Full-task]{
The advisor is cut off before completing the substitution, and the SLM resumes from that setup, and obtains the final answer $19$.
The advisor supplies the structural reduction, and the SLM performs the remaining algebra and final answer extraction.
}

\end{generationexample}


\begin{generationexample}{
    MATH500, Problem 288
    \hfill
    \textnormal{Advisor budget: 512 tokens}
}
\begin{problembox}
One line is defined by
\[
\begin{pmatrix}
3 \\
-10 \\
1
\end{pmatrix}
+
t
\begin{pmatrix}
2 \\
-9 \\
-2
\end{pmatrix}.
\]

Another line is defined by
\[
\begin{pmatrix}
-5 \\
-3 \\
6
\end{pmatrix}
+
u
\begin{pmatrix}
4 \\
-18 \\
-4
\end{pmatrix}.
\]

These two lines are parallel. Find the distance between these two lines. Let's think step by step and output the final answer within \(\boxed{}\).
\end{problembox}
To find the distance between two parallel lines in 3D space, we can use
the formula involving the cross product of the direction vectors of the
lines and the vector connecting a point on each line. The distance $d$
between two parallel lines given by
$\mathbf r_1=\mathbf a_1+t\mathbf b_1$ and
$\mathbf r_2=\mathbf a_2+u\mathbf b_2$ is given by
\[
d=
\frac{|(\mathbf b_1\times\mathbf b_2)\cdot
(\mathbf a_2-\mathbf a_1)|}
{|\mathbf b_1\times\mathbf b_2|}.
\]

Here, the direction vectors are
\[
\mathbf b_1=
\begin{pmatrix}
2\\-9\\-2
\end{pmatrix},
\qquad
\mathbf b_2=
\begin{pmatrix}
4\\-18\\-4
\end{pmatrix}.
\]
We can see that $\mathbf b_2=2\mathbf b_1$, so the lines are indeed
parallel. The vector connecting a point on the first line to a point on
the second line is
\[
\mathbf a_2-\mathbf a_1
=
\begin{pmatrix}
-5\\-3\\6
\end{pmatrix}
-
\begin{pmatrix}
3\\-10\\1
\end{pmatrix}
=
\begin{pmatrix}
-8\\7\\5
\end{pmatrix}.
\]

Next, we need to compute the cross product
$\mathbf b_1\times\mathbf b_2$:
\[
\mathbf b_1\times\mathbf b_2
=
\begin{vmatrix}
\mathbf i & \mathbf j & \mathbf k\\
2 & -9 & -2\\
4 & -18 & -4
\end{vmatrix}
=
\begin{pmatrix}
0\\0\\0
\end{pmatrix}.
\]

Since $\mathbf b_2=2\mathbf b_1$, the cross product
$\mathbf b_1\times\mathbf b_2$ is the zero vector. This means the
denominator in the distance formula is zero, which indicates that the
lines are actually the same line, and the distance between them is
zero. However, if the lines were truly parallel but not the same, we
would need to use the non-zero cross product to find the distance.

Given the zero cross product, we need to re-evaluate the problem. Since
the lines are parallel, we can use the advisor to get the specific
calculation needed to confirm the distance.

\begin{qmakerbox}
\qmakerrequest{
Two lines in 3D space are defined as follows:
\[
\mathbf r_1=
\begin{pmatrix}
3\\-10\\1
\end{pmatrix}
+t
\begin{pmatrix}
2\\-9\\-2
\end{pmatrix}
\]
and
\[
\mathbf r_2=
\begin{pmatrix}
-5\\-3\\6
\end{pmatrix}
+u
\begin{pmatrix}
4\\-18\\-4
\end{pmatrix}.
\]
It is given that these lines are parallel. How do you calculate the
distance between these two parallel lines?
}

\qmakerbudget{
Answer in 5--7 lines.
}
\end{qmakerbox}

\begin{advisorbox}
First, observe that the direction vector of the second line is exactly
twice the first, confirming they are parallel. To find the distance,
select a point $P$ on the first line (e.g., $t=0$) and a point $Q$ on
the second line (e.g., $u=0$). Calculate the vector $\vec{PQ}$
connecting these two points. The distance is the magnitude of the
component of $\vec{PQ}$ perpendicular to the direction vector
$\mathbf d$, found using the formula
\[
\frac{\|\vec{PQ}\times\mathbf d\|}{\|\mathbf d\|}.
\]
Compute the cross product of $\vec{PQ}$ and $\mathbf d$, find its
magnitude, and divide by the magnitude of $\mathbf d$. This yields the
shortest distance between the two parallel lines.
\end{advisorbox}

Based on the advisor's response, we need to follow these steps to find
the distance between the two parallel lines:

\begin{enumerate}
    \item Identify a point on each line. For the first line, when
    $t=0$, the point is
    $\mathbf a_1=(3,-10,1)^T$. For the second line, when $u=0$, the
    point is $\mathbf a_2=(-5,-3,6)^T$.
    \item Calculate
    \[
    \vec{PQ}
    =\mathbf a_2-\mathbf a_1
    =
    \begin{pmatrix}
    -8\\7\\5
    \end{pmatrix}.
    \]
    \item Use the direction vector
    $\mathbf d=(2,-9,-2)^T$ to find
    $\vec{PQ}\times\mathbf d$.
    \item Compute the magnitude of the cross product
    $\|\vec{PQ}\times\mathbf d\|$.
    \item Divide the magnitude of the cross product by
    $\|\mathbf d\|$.
\end{enumerate}

Let's perform these calculations step by step.

\par\medskip
\noindent\textit{[... Omitted for brevity ...]}
\par\medskip

Thus, the distance between the two lines is
\[
\boxed{7}.
\]

\interpretation[Transform]{
The SLM initially attempts to use an inappropriate cross-product formula for the distance between parallel lines and incorrectly interprets the zero cross product of the direction vectors.
The advisor redirects the SLM to the correct point-to-line distance formula using a connecting vector and a single direction vector. The SLM then carries out the numerical computation independently and obtains the final answer $7$.
}

\end{generationexample}


\begin{generationexample}{
    MATH500, Problem 287
    \hfill
    \textnormal{Advisor budget: 512 tokens}
}
\begin{problembox}
One gear turns $33\frac{1}{3}$ times in a minute. Another gear turns
$45$ times in a minute. Initially, a mark on each gear is pointing due
north. After how many seconds will the two gears next have both their
marks pointing due north?
\end{problembox}
To determine after how many seconds the two gears will next have both
their marks pointing due north, we need to find the least common
multiple (LCM) of the number of turns each gear makes in a minute and
then convert that time into seconds.

First, let's convert the number of turns per minute for each gear into
a more manageable form. The first gear turns $33\frac13$ times in a
minute, which can be written as $\frac{100}{3}$ turns per minute. The
second gear turns $45$ times in a minute.

The LCM of $\frac{100}{3}$ and $45$ will give us the number of turns
per minute after which both gears will be at the same position relative
to their starting points. To find the LCM, we need to find the LCM of
the numerators when the fractions are expressed with a common
denominator.

The first gear's turns per minute is $\frac{100}{3}$, and the second
gear's turns per minute is $\frac{135}{3}$, since
$45=\frac{135}{3}$. The LCM of $100$ and $135$ will be the numerator
of the LCM of the fractions.

The prime factorizations are
\[
100=2^2\times5^2
\]
and
\[
135=3^3\times5.
\]

The LCM of $100$ and $135$ is
\[
2^2\times3^3\times5^2=1800.
\]

So, the LCM of $\frac{100}{3}$ and $\frac{135}{3}$ is
$\frac{1800}{3}=600$ turns per minute. This means that after 600 turns,
both gears will be at the same position relative to their starting
points.

Since there are 60 seconds in a minute, the time in seconds is
\[
600\text{ turns}\times
\frac{1\text{ minute}}{60\text{ seconds}}
=
10\text{ minutes}
=
10\times60
=
600\text{ seconds}.
\]

However, we need to confirm that this is the smallest such time. To do
this, we can use the advisor to verify the calculation and ensure we
have the correct LCM.

\begin{qmakerbox}
\qmakerrequest{
Two gears turn $33\frac13$ times and $45$ times in a minute,
respectively. Initially, a mark on each gear is pointing due north.
After how many seconds will the two gears next have both their marks
pointing due north?
}

\qmakerbudget{
Answer in 5--7 lines.
}
\end{qmakerbox}

\begin{advisorbox}
First, convert the rotation rates to revolutions per second: Gear A
turns $33\frac13=\frac{100}{3}$ times per minute, which is
$\frac{100}{180}=\frac59$ revolutions per second. Gear B turns 45
times per minute, which is $\frac{45}{60}=\frac34$ revolutions per
second.

For both marks to point north again, the elapsed time must be a common
multiple of the time each gear takes to complete one full revolution.
The period for Gear A is $\frac95$ seconds, and for Gear B is
$\frac43$ seconds.

We need the least common multiple of these two fractions:
\[
\operatorname{LCM}\left(\frac95,\frac43\right)
=
\frac{\operatorname{LCM}(9,4)}
{\operatorname{GCD}(5,3)}.
\]
The LCM of the numerators 9 and 4 is 36, and the GCD of the
denominators 5 and 3 is 1. Thus, the time required is
\[
\frac{36}{1}=36
\]
seconds. This is the first time both gears simultaneously return their
marks to the north position.
\end{advisorbox}

The advisor has confirmed that the least common multiple of the periods
of the two gears is 36 seconds. Therefore, the two gears will next have
both their marks pointing due north after
\[
\boxed{36}
\]
seconds.

\interpretation[Full-task]{
The QMaker does not isolate a missing subproblem or request a specific piece of guidance. 
It simply restates the original benchmark problem and asks for its answer. 
The advisor then solves the task end to end.
The SLM performs no additional derivation after the advisor call and merely repeats the advisor's conclusion.
The efficiency gain in this case solely comes from the budget instruction. 
}

\end{generationexample}

\begin{generationexample}{
    LiveCodeBench v6, Problem 236
    \hfill
    \textnormal{Advisor budget: 1024 tokens}
}
\begin{problembox}
You are an expert Python programmer. You will be given a question (problem specification) and will generate a correct Python program that matches the specification and passes all tests.

Below is the question:

There are N cities in a certain country.
You will travel from your office in city 1 to a destination in city N, via zero or more cities.
Two types of transportation are available: company car and train. The time required to travel from city i to city j is as follows:

- $D_({i,j}) \times A$ minutes by company car, and
- $D\_({i,j}) \times B + C$ minutes by train.

You can switch from company car to train, but not vice versa.
You can do so without spending time, but only in a city.
What is the minimum time in minutes to travel from city 1 to city N?

\par\medskip
\textit{[...Constraints, Input, Output, Examples, ...]}
\par\medskip

\end{problembox}
\studentturntitle{Student turn 1}

\begin{studentcode}
import sys
from typing import List

def read_input() -> List[List[int]]:
    n, a, b, c = map(int, sys.stdin.readline().split())
    d = []
    for _ in range(n):
        d.append(list(map(int, sys.stdin.readline().split())))
    return n, a, b, c, d

def min_time_to_travel(n: int, a: int, b: int, c: int, d: List[List[int]]) -> int:
    # Initialize the distance matrix
    dist = [[float('inf')] * n for _ in range(n)]
    for i in range(n):
        dist[i][i] = 0
    
    # Fill the distance matrix using the given distances
    for i in range(n):
        for j in range(n):
            if d[i][j] > 0:
                dist[i][j] = min(d[i][j] * a, d[i][j] * b + c)
    
    # Floyd-Warshall algorithm to find the shortest path
    for k in range(n):
        for i in range(n):
            for j in range(n):
                dist[i][j] = min(dist[i][j], dist[i][k] + dist[k][j])
    
    return dist[0][n-1]

if __name__ == "__main__":
    n, a, b, c, d = read_input()
    result = min_time_to_travel(n, a, b, c, d)
    print(result)
\end{studentcode}

\begin{qmakerbox}
\qmakerrequest{
Given:
- \( N \) cities in a country.
- You start in city 1 and want to travel to city \( N \).
- Two types of transportation are available: company car and train.
- The time required to travel from city \( i \) to city \( j \) is:
  - \( D_{i,j} \times A \) minutes by company car, and
  - \( D_{i,j} \times B + C \) minutes by train.
- You can switch from company car to train, but not vice versa.
- You can switch without spending extra time, but only in a city.

How would you compute the minimum time in minutes to travel from city 1 to city \( N \)?
}

\qmakerbudget{
20 lines
}
\end{qmakerbox}

\begin{advisorbox}
1. Define $D_{i,j}$ as the distance between city $i$ and city $j$.

2. Let $T_{car}(i,j) = D_{i,j} \times A$ and $T_{train}(i,j) = D_{i,j} \times B + C$.

3. Since switching is allowed only once (car to train), split the path into two segments.

4. Segment 1: Travel from city 1 to an intermediate city $k$ using only the company car.

5. Segment 2: Travel from city $k$ to city $N$ using only the train.

6. The total time for a specific switch point $k$ is $Time(k) = \min_{path_1} \sum T_{car} + \min_{path_2} \sum T_{train}$.

7. Compute the shortest path from city 1 to all cities $k$ using only car edges (Dijkstra).

\par\medskip
\textit{[... Omitted for brevity ...]}
\par\medskip

19. The constraint "switch only once" is satisfied by iterating over every possible switch city $k$.

20. Return the computed minimum time as the solution.
\end{advisorbox}

\studentturntitle{Student turn 2}

\begin{studentcode}
import sys
from typing import List

def read_input() -> List[List[int]]:
    n, a, b, c = map(int, sys.stdin.readline().split())
    d = []
    for _ in range(n):
        d.append(list(map(int, sys.stdin.readline().split())))
    return n, a, b, c, d

def min_time_to_travel(n: int, a: int, b: int, c: int, d: List[List[int]]) -> int:
    # Initialize the distance matrix
    dist_car = [[float('inf')] * n for _ in range(n)]
    dist_train = [[float('inf')] * n for _ in range(n)]
    
    # Fill the distance matrix using the given distances
    for i in range(n):
        for j in range(n):
            if d[i][j] > 0:
                dist_car[i][j] = d[i][j] * a
                dist_train[i][j] = d[i][j] * b + c
    
    # Compute the shortest path from city 1 to all cities using only car edges
    for k in range(n):
        for i in range(n):
            for j in range(n):
                dist_car[i][j] = min(dist_car[i][j], dist_car[i][k] + dist_car[k][j])
    
    # Compute the shortest path from all cities to city N using only train edges
    for k in range(n):
        for i in range(n):
            for j in range(n):
                dist_train[i][j] = min(dist_train[i][j], dist_train[i][k] + dist_train[k][j])
    
    # Find the minimum time by considering all possible switch points
    min_time = float('inf')
    for k in range(n):
        if k != 0 and k != n-1:
            time = dist_car[0][k] + dist_train[k][n-1]
            min_time = min(min_time, time)
        elif k == 0:
            time = dist_train[0][n-1]
            min_time = min(min_time, time)
        elif k == n-1:
            time = dist_car[0][n-1]
            min_time = min(min_time, time)
    
    return min_time

if __name__ == "__main__":
    n, a, b, c, d = read_input()
    result = min_time_to_travel(n, a, b, c, d)
    print(result)
\end{studentcode}

\interpretation[Full-task]{
    The advisor provides the solution strategy entirely in natural-language steps, describing how to separate the car and train shortest-path computations around a single switch city.
    However, the SLM fails to faithfully integrate this guidance into the final code.
    The student utilizes Floyd--Warshall computation, instead of suggested Dijstra-based approach.
    This example illustrates one example of integration failure by the SLM.
}

\end{generationexample}